\documentclass[11pt]{article}

\usepackage[final]{acl} 
\usepackage{times}
\usepackage{latexsym}
\usepackage[T1]{fontenc}
\usepackage[utf8]{inputenc}
\usepackage{microtype}
\usepackage{inconsolata}
\usepackage{graphicx}

\usepackage{booktabs}
\usepackage{array}
\usepackage{amsmath}
\usepackage{textcomp}
\usepackage{pifont}
\usepackage{tabularx}

\usepackage{multirow}
\usepackage{paralist}
\usepackage{verbatim}
\usepackage{xcolor}

\newcommand{\argsup}[2]{[#1]$^{\mbox{\textsc{#2}}}$}
\newcommand{\sbj}[1]{\argsup{#1}{sbj}}
\newcommand{\obj}[1]{\argsup{#1}{obj}}
\newcommand{\sbert}{SBERT}

\newcommand{\tacred}{FS-TACRED}
\newcommand{\fewrel}{FS-FewRel}
\newcommand{\qwensmall}{Qwen3-4B}
\newcommand{\qwenlarge}{Qwen3-14B}
\newcommand{\gemmasmall}{Gemma3-4B}
\newcommand{\gemmalarge}{Gemma3-12B}
\newcommand{\sm}{SoftMatcher}
\newcommand{\norel}{\begin{small} \texttt{no\_relation} \end{small}}

\usepackage{todonotes}

\title{Structured Semantic Information Helps Retrieve Better Examples\\for In-Context Learning Applied to Few-Shot Relation Extraction}

\author{Aunabil Chakma, Mihai Surdeanu, and Eduardo Blanco\\
  University of Arizona, Tucson, AZ, USA \\
  \texttt{\{aunabilchakma, msurdeanu, eduardoblanco\}@arizona.edu} \\}

\begin{document}

\maketitle

\begin{abstract}
This paper presents several strategies to automatically obtain
additional examples for in-context learning, effectively transforming relation extraction from a $1$-shot to a few-shot setting.
Specifically, we introduce a novel strategy for example selection, 
in which new examples are selected based on the similarity of their underlying syntactic-semantic structure to the provided $1$-shot example. 
We show that our strategy results in complementary word choices and sentence structures compared to LLM-generated examples. 
When both strategies are combined, the resulting hybrid system achieves a more holistic picture of the relations of interest than either method alone.
Our framework transfers well across datasets (\tacred{} and \fewrel{}) and LLM families (Qwen and Gemma).
Overall, our hybrid system consistently outperforms alternative strategies 
achieving state-of-the-art performance on \tacred\ and strong gains on a customized FewRel subset.

\end{abstract}


\section{Introduction}
\label{sec:introduction}

Relation extraction (RE) is a core NLP task that 
identifies relations between entity pairs~(e.g., the place of birth or employer of a person).
Extracting relations is often a component of other tasks such as knowledge graph construction~\cite{schneider-etal-2022-decade}.
In turn, knowledge graphs coupled with retrieval-augmented approaches have been shown
to improve LLM reasoning~\cite{sui-etal-2025-fidelis} and
reduce hallucinations~\cite{wagner-etal-2025-mitigating},
thus potentially benefiting any end-user application including question answering~\cite{Xu_Li_Zhang_Lin_Zhu_Zheng_Wu_Zhao_Xu_Chen_2025}.

Traditional relation extraction approaches require many training examples per relation~\cite{marsh-1998-tipster}.
More recently, few-shot relation extraction (FSRE) has received substantial attention~\cite{fs-tacred,FSNYT}.
In this setting, relations must be extracted given nothing but the relation name, a brief description, and a handful of examples.
Because of this challenging setting,
previous work obtains relatively low results (e.g., $\approx$24~F1  on \tacred{}).
Existing approaches to FSRE
include those grounded in prototypes~\cite{RAPL_prototype_appraoch1,Relation_description_prototype_appraoch2}
and typically require training or fine-tuning~(Section~\ref{sec:related_work}).



\begin{figure*}
\centering
\includegraphics[width=0.9\textwidth]{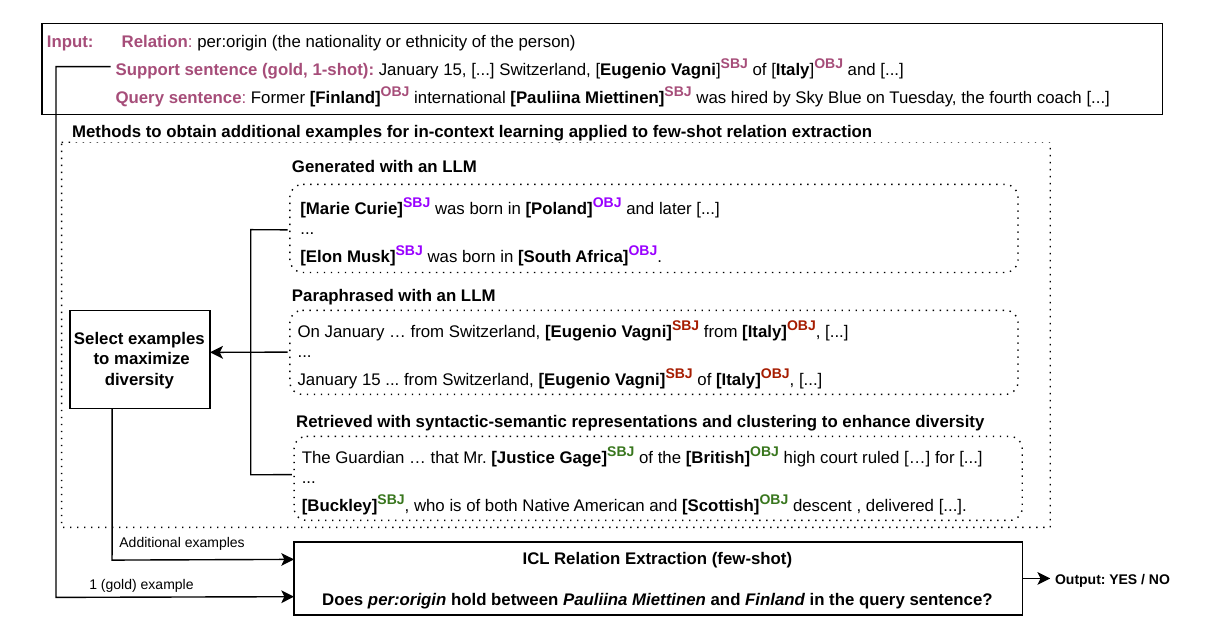} 
\caption{
Our approach to in-context learning for $1$-shot relation extraction.
Our method complements the (gold) $1$-shot ("Support sentence" in the figure) with 4 or 9 additional examples, turning the setup into a few-shot setting ($5$- or $10$-shot), 
generated using shallower LLM-based methods (i.e., paraphrasing the gold example, or generating new examples based on the relation definition) or deeper methods using the underlying syntactic-semantic representations of candidate examples.
When combining these strategies, the resulting few-shot relation extraction system achieves better results across two corpora (\tacred\ and \fewrel) and two language models (Qwen3 and Gemma3). 
}
\label{fig:overview_method}
\end{figure*}

In-context learning (ICL) with LLMs~\citep{llm_are_few_shot_learner}
for FSRE is a natural choice yet how additional examples and diversity affect performance remains underexplored. In particular, the best strategy for example selection in FSRE remains unknown.
To address this gap, in this paper we explore\footnote{Code and data are available at \url{https://github.com/clulab/example-selection-icl-fsre}.} several strategies to automatically obtain additional examples for the relation of interest.
In particular, we explore two main strategies: using 
LLMs to generate new examples and retrieving examples from unannotated corpora {\em based on their deeper syntactic-semantic similarity} with the provided $1$-shot example.

Figure \ref{fig:overview_method} presents our strategies to obtain additional examples for $1$-shot relation extraction.
We automatically combine the $1$-shot gold example with additional examples that are: (a)~generated with LLMs or (b)~retrieved from unannotated corpora using their syntactic-semantic similarity with the gold example.
For LLM-based generation, we either paraphrase the gold example while preserving the subject, object, and relation, 
or generate new examples conditioned on the relation name and its natural language definition, encouraging alternative lexicalizations and sentence structures.
Our key novelty is the use of lexico-syntactic rules in a semantic space for example retrieval.
These structured rules capture how the subject and object are connected in a sentence through their core syntactic structure and semantic context (Section~\ref{subsec:retrieve-examples-from-corpora}).
Importantly, the semantic representations of these rules are learned in a self-supervised fashion that is agnostic to the relations of interest, similar to word embeddings~\cite{10.5555/2999792.2999959}.
Further, we emphasize diversity in the retrieved examples by: (a)
clustering candidates using their semantic representations and selecting one representative example per cluster, and (b) using an LLM to select the most diverse examples from a pool of example candidates created with the above strategies.


The main contributions of this paper are:
  
  
 
\begin{itemize}
  \item Several strategies to obtain additional examples for FSRE that range from shallower (e.g., LLM paraphrasing) to deeper (i.e., using representations of the syntactic-semantic structures of both the $1$-shot gold example and the candidates themselves).
  \item Methods to emphasize example diversity (i.e., diverse word choices and sentence structures) from a pool of candidate examples constructed using the above strategies. We investigate two directions: (a) explicit clustering of examples and selecting one representative example per cluster, and (b) prompt engineering.
  \item Demonstrating that a hybrid approach balancing similarity and diversity obtains new state-of-the-art (SOTA) performance for FS-TACRED, especially with smaller language models, outperforming prior methods.
\end{itemize}

\section{Related Work}
\label{sec:related_work}

Early approaches and datasets for RE
targeted supervised learning~\cite{marsh-1998-tipster,doddington-etal-2004-automatic}.
These approaches are effective when trained on hundreds or thousands of examples per relation~\cite{end2end_supervised_work1,svm_supervised_work2}.
However, they rarely generalize across domains (i.e., same relation types expressed in documents from different sources)
and can only identify at inference time the relations they were trained on.
Distant supervision~\cite{mintz-etal-2009-distant} has been explored to eliminate manual annotations,
but other limitations remain.
 

\paragraph{Few-Shot Relation Extraction}
Most modern FSRE methods bypass the need to know the relations to be extracted at inference time and avoid manual annotations beyond a handful of examples per relation.
Representative approaches include prototype-based classification
using aggregated relation representations~\citep{RAPL_prototype_appraoch1,Relation_description_prototype_appraoch2, prototype_10, prototype_11, prototype_12}.  
Embedding-based nearest-neighbor methods are also widely used~\citep{fs-tacred, GPT-RE, embedding_work2}. 
While effective, these methods lack flexibility under relation or domain shifts. 
In contrast, neuro-symbolic approaches emphasize explicit relational cues~\citep{softmatcher,anchorword}, which improves their in- and out-of-domain performance.
These properties motivate our approach to retrieve examples grounded in syntactic-semantic representations.
Moreover, most existing work on FSRE is evaluated on TACRED~\citep{TACRED}, FewRel~\citep{fewrel}, and NYT~\citep{NYT}. 
In this work, we focus on a more challenging variant introduced by~\citet{fs-tacred}, 
which contains a large proportion of NOTA (none-of-the-above) instances, making the task substantially harder (Section~\ref{sec:experiments}).

While we are not the first to use in-context learning and LLMs for FSRE~\citep{COT-RE,Unleash-ICL}, 
%
we are the first to use syntactic-semantic similarity to retrieve examples for ICL applied to FSRE. Our strategies for example selection improve both similarity and diversity. 
Moreover, we demonstrate that our approach transfers across benchmarks and LLMs, using the same prompt for each component across all models.\footnote{
  We do not compare our results with~\citet{COT-RE} and~\citet{Unleash-ICL}
  because they work with customized few-shot versions of TACRED and FewRel for which the exact few-shot evaluation setups are not publicly available. 
}

\paragraph{Example selection for In-Context Learning}
Prior work has studied example selection
for ICL applied to several NLP tasks including FSRE.  
Existing approaches mainly rely on embedding-based similarity~\citep{GPT-RE,C-ICL}
or learning-based selection~\citep{RetICL,ITERR,RetrieveICL} using task supervision, model feedback~\citep{LENS,RetrieveICL,ITERR}, and diversity-aware criteria~\citep{LENS,ICL_with_itr_dem_selection}.
In contrast, we leverage syntactic-semantic representations to guide example retrieval, 
an under-explored line of research in FSRE. 
LLM-based data generation techniques are also explored for RE to improve training and prediction~\citep{LLM_GEN_examples1, LLM_GEN_examples2}.  
We explore LLM-based example generation as a simple baseline and also incorporate it into our hybrid method.

\paragraph{Similarity and Diversity in Relation Extraction}
Diversity plays a crucial role in in-context learning (ICL)~\citep{rolediversityincontextlearning, LLM_GEN_examples1, diverse_paper_1, diverse_paper_2}. 
However, most prior work measures diversity using sentence-level embeddings, which often capture irrelevant context and surface variation rather than the differences in the relation holding between two entities. 
In contrast, we (a) focus on the minimal context needed to express a relation by embedding the shortest path between entity pairs in the dependency-tree, 
 and (b) enforcing diversity directly at the rule level, yielding a more informative similarity-diversity trade-off for relation extraction.

\section{The Few-Shot Relation Extraction Task}
\label{sec:few-shot-relation-extraction-task}
Few-shot relation extraction is formulated as an episodic $n$-way $k$-shot classification task.  
A relation holds between two entities: the {\em subject} and {\em object}.
Each episode consists of a $k$-shot learning instance defined by
(a) a set of candidate relations $\mathcal{R} = \{r_1, \ldots, r_n\}$ along with $k$ support sentences exemplifying each relation
and
(b) a query sentence with a subject and object named entities tagged.
The task is to predict which of the $n$ relations hold between the entities in the query sentence, or output \norel{} if none apply. 
We target the standard $5$-way $1$-shot setting as defined in existing corpora (Section \ref{sec:experiments}).
In particular, we explore in-context learning and
strategies to complement the single gold example per relation with automatically obtained additional examples.



\paragraph{In-Context Learning for Few-Shot Relation Extraction}
\label{sec:relation-wise-icl-inference}

We reformulate the $5$-way $1$-shot relation extraction task as five independent $1$-shot binary classifications.
Each classification task predicts whether a relation holds between the subject and object in the query sentence
(see few-shot prompt in Figure~\ref{t:prompt_one-relation-inference}).
We output \norel{} if none hold,
and a random relation out of all that hold if more than one is predicted.\footnote{This scenario is very unlikely: 0.26\% of query sentences.}
Additionally, prior to in-context learning,
we include a filter that enforces compatibility between the named entity types in the query sentence and the type constraints of the target relation. 
We use in-context learning (Figure~\ref{t:prompt-entity-resolution}) to match these named entity types.
This filter was added because it substantially 
reduces compute ($86.04$\% of candidate relations are discarded), 
and yields empirical benefits (Section \ref{sec:qualitative_analysis}).

We also experimented with in-context learning to decide which 
of the five candidate relations hold between the entities in the query sentence using a single LLM call, 
i.e., where the LLM functions as a multi-class classifier.
We discarded it after observing consistently lower results than with the binary approach outlined above (Figure~\ref{t:prompt_multi-relation-inference}).

\section{From $1$-shot Relation Extraction to Many-Shot Relation Extraction}
\label{sec:example-selection-fsre}

In this section, we propose strategies for automatically transforming the $1$-shot relation extraction task
into a many-shot task, where the additional examples are automatically obtained.
We consider LLM-based strategies
as well as retrieval-based strategies from unannotated corpora.
Table \ref{t:diversity_examples} illustrates the additional examples obtained with all the strategies.
As we shall see, a hybrid approach that balances similarity to the $1$-shot example with contextual diversity is the most effective.


\subsection{Obtaining Examples with LLMs}

Our first strategies use LLMs to obtain additional examples and are arguably the simplest ones.
 
\noindent
\textbf{Paraphrasing with LLMs:}
We paraphrase the original (gold) support sentence for a relation with the prompt in Figure~\ref{t:prompt-paraphrase}.
The prompt is designed to ensure that
(a) the same relation holds between the same named entities
and
(b) the subject and object entities remain tagged.

\noindent
\textbf{Generating with LLMs:}
We also explore a $1$-shot example generation strategy with the prompt in Figure~\ref{t:prompt-new-example-generation}.
This prompt uses the original (gold) support sentence as an example,
and is designed to encourage different wordings and sentence structures while preserving the target relation.
As we'll see in Section~\ref{sec:qualitative_analysis}, the LLMs tend to generate examples using their parametric knowledge (e.g., relations involving celebrities or politicians). 

\subsection{Retrieving Examples from Unannotated Corpora Using Syntactic-Semantic Representations} 
\label{subsec:retrieve-examples-from-corpora}

Our alternative to obtaining examples with LLMs is to retrieve them from large unannotated corpora.
This approach has two advantages.
First, retrieved examples reflect natural language usage patterns and can expose in-context learning to more realistic examples than LLM-synthesized examples.
Second, retrieval can leverage precomputed representations to scale to millions of candidate sentences, 
enabling us to search for examples that are both semantically relevant and structurally diverse.


We retrieve examples from a subset of UMBC WebBase~\citep{umbc},
a corpus of over 3 billion words collected from a web crawl of 100 million web pages from more than 50,000 websites~\cite{10.1016/S1389-1286(00)00063-3}.
Specifically, we use the subset created by~\citet{softmatcher}.
After removing duplicates and downsampling to ensure uniform frequency of named entity types,
this subset contains 2.3M sentences
containing two named entities belonging to types compatible with the corpora we work with (Section \ref{sec:experiments}).
Prior work has used this UMBC subset as a source of self-supervised training~\citep{softmatcher}. 
In contrast, we use this dataset to retrieve examples for ICL.
The retrieval strategies always select examples based on cosine similarity.
The main differences are
(a) the representations for the gold support sentence and candidate sentences from UMBC (\sbert{} or syntactic-semantic rule representations)
and
(b) the methods to select candidate examples out of the closest ones.

\paragraph{Sentence Representations from \sbert{}}
We use \sbert{}~\citep{sentencebert} as a baseline.
Note that these representations are meant to embed the meaning of a sentence
without emphasizing the subject and object named entities,
the relation between them,
or the context within the sentence that is most indicative of the relation.

\paragraph{Syntactic-Semantic Rule Representations}
In the real-world, it is often the case that a relation holds between two entities in a sentence, yet
(a)~only a portion of a sentence signals the relation between subject and object entities,
and
(b)~context other than the tokens between subject and object entities are indicative of which relation holds (see examples in Table \ref{t:diversity_examples}).
Our rationale behind using syntactic-semantic representations is to pinpoint and leverage the specific context
that is most beneficial for FSRE.

In particular, by lexico‑syntactic rules we mean structured representations that encode the shortest syntactic dependency path between subject and object entities, 
combining lexical items, dependency relations, their directionality, and the named entity types of the subject and object entities.
These paths are extracted from dependency parses produced by tools such as Stanford CoreNLP\footnote{\url{https://stanfordnlp.github.io/CoreNLP/}} or Odinson~\cite{odinson}, 
and are used in prior work~\cite{softmatcher} to capture relational structure.

Semantic rule representations are created using the 
lexico-syntactic rule extractor and semantic rule matcher of~\citet{softmatcher}.
The rule matcher, named \sm{}, was trained following a self-supervised approach
and is used in two main stages during retrieval.
First, semantic representations embedding the lexico-syntactic rule---not the sentence they originate from---are obtained
for the (gold) support sentence and candidate sentences containing the same named entity types.
Second, candidate sentences are selected based on cosine similarity between the rule representation
of the support sentence and candidate sentences.
The second step is facilitated at scale by a FAISS index~\cite{faiss} accounting for the tokens in a sentence, named entity types, and rule representation.


Consider the following support sentence for the \emph{per:origin} relation: 
``Just this week we have the \obj{European} G6 group, lead by the UK's Home Secretary \sbj{John Reid}, promising to work together to make the internet a~[...].''
The lexico-syntactic rule is as follows:
{\tt [entity=NATIONALITY]+ <amod group >acl:relcl lead >nmod\_by [entity=PERSON]+},
which can be read as
``a nationality named entity
  modifying (incoming syntactic dependency `amod') \emph{group},
  which is modified by an outgoing dependency acl:relcl headed by \emph{lead}, 
  which is modified by (outgoing dependency `nmod' headed by \emph{by}) a person named entity.''
After obtaining semantic rule representations, the \sm{}
finds sentences with
(a)~the same named entity types and
(b)~different contexts expressing the same relation.
For example, 
``\sbj{Buckley}, who is of both Native American and \obj{Scottish} descent, delivered the sermon [...]''
is the closest sentence obtained by rule representations.
  

\paragraph{Enhancing Diversity}

Selecting the most similar examples often yields redundant examples that express the relation
in nearly identical ways~(same entities, very similar word choices indicating the relation, etc.).
To encourage diversity, we explore strategies to ensure differences among the selected examples.
To this end, we first select the $m$ candidate sentences above a similarity threshold $\tau$ tuned with validation instances.
Second, we use $k$-means or $k$-means++ to obtain $k=\lfloor \sqrt{m} \rfloor$ clusters.
Third, we select as many clusters as additional examples are desired by choosing
(a) randomly,
(b) the clusters closest to the support sentence,
or
(c) the clusters that maximize pairwise intra-cluster distances after initially selecting the furthest cluster to the support sentence.
Fourth, we select the closest sentence to each cluster centroid as a representative example.
In total, we select examples following four diversity-oriented strategies ($k$-means with random or closest cluster selection, and $k$-means++ with farthest or closest cluster selection), which will be empirically validated (Section \ref{sec:results}).


\subsection{Hybrid Example Selection}
Obtaining examples with LLMs and retrieving them from unannotated corpora offer complementary strengths.
The former tend to have a very similar form to the original support sentence,
while the latter represent more natural usage and diversity.
We therefore explore a hybrid strategy that combines examples from both sources.
Specifically, we shuffle the examples from both sources and select the desired amount of examples using the prompt in Figure~\ref{t:prompt-pick_examples}.
Intuitively, the hybrid approach balances the presence of
(a) similar examples to the support sentences~(obtained with LLMs)
and
(b) diverse examples with respect to the support sentence~(retrieved from unannotated corpora).

\subsection{Summarizing Examples}
Prior work~\citep{summarization-help} has shown that summarizing examples can improve in-context learning 
by reducing spurious content to the minimal content needed for the task at hand.
In relation extraction, summarization would simplify examples to contain only the information needed to extract the relation of interest between
the subject and object named entities.
For example, \emph{both Native American} and all tokens after \emph{descent}
could be deleted without altering the signal for the relation between
\emph{Buckley} and \emph{Scottish} in
``\sbj{Buckley}, who is of both Native American and \obj{Scottish} descent, delivered the sermon during Saturday evening's workship service [...].''
We experiment both with the examples
(a) generated and retrieved
and
(b) their summarized versions with the prompt in Figure~\ref{t:prompt_summariztation}.

\section{Experimental Setup}
\label{sec:experiments}

We experiment primarily with two datasets (\tacred\ and \fewrel)
and four LLMs (two sizes of two LLM families)
as described below.

\paragraph{Datasets}
Our primary evaluation benchmark is \tacred{}~\cite{fs-tacred}, a few-shot reformulation of TACRED~\citep{TACRED}.
We use the five test episode files (all episodes are $5$-way $1$-shot) released by \tacred{}.
Each file has 10,000 episodes and 3 queries per episode.
We do not train any model on these episodes.
Instead, we use the development episodes as held-out data to tune hyperparameters (see Section~\ref{subsec:hyperparameters-retrieval} for details).

To test whether our strategies transfer across benchmarks,
we also experiment with \fewrel{}, a few-shot version of FewRel~\cite{fewrel} we created following the algorithm by~\citet{FSNYT}.
Because the labels for the test split are not publicly available, we report results with the development split.
Specifically, we work with the six relations whose subject and object entities belong to named entity types compatible with the relation types in \tacred{}.
Similar to \tacred{}, we construct 5 sets of 10,000 $5$-way $1$-shot episodes, including a \texttt{no\_relation} label,
and each episode has three query sentences.

The majority label in both \tacred{}\ and \fewrel{}\ is \norel{} (97\% and 95\%).
This is by design,
as realistic few-shot relation extraction requires models to account for the most common case:~none of the five relations hold between the entities in the query sentence.

We chose not to report results on FS-NYT~\cite{FSNYT} because it contains primarily relations between well-known named entities and locations
(e.g., \emph{place of birth} of famous people), which were very likely seen during the LLM pretraining.
As a result, the vast parametric knowledge in pre-trained LLMs is sufficient to obtain much better results than with \tacred{} and \fewrel{}
with a prompt that provides the subject and object entities \emph{and no additional context from the query sentence}.
We refer the reader to Table \ref{t:parametric-test} for the results to check for parametric knowledge within the three datasets.

\paragraph{LLMs and Implementation Details}
We design the pipeline to rely on a single LLM across all components~(i.e., NER filtering, LLM-based generation and paraphrasing of examples, hybrid example selection, etc.), 
ensuring consistency, and simplifying deployment and transferability across different model families.
Specifically, we evaluate the full system with four LLMs available in HuggingFace~\cite{wolf-etal-2020-transformers}:
two sizes of two LLM families (\qwensmall{}, \qwenlarge{}, \gemmasmall{}, and \gemmalarge{}).
Further details on decoding and inference are provided in Section~\ref{subsec:decoding_details}, 
while hyperparameter tuning and retrieval efficiency are described in Section~\ref{subsec:hyperparameters-retrieval}.


\paragraph{Baselines and Evaluation Metric}
In addition to previous work on \tacred{} ($5$-way, $1$-shot setting),
we consider a $1$-shot ICL baseline that considers the original support sentence as the only example.
We compare this baseline against $1$+$4$- and $1$+$9$- shot ICL approach,
where one example is the original support sentence
and the other four or nine are obtained with the strategies described in Section \ref{sec:example-selection-fsre}.
Following previous work, we report the average Precision, Recall, and F1 scores,
along with their standard deviations and statistical significance results.
Following previous work, we calculate metrics excluding \norel{}.

\section{Results}
\label{sec:results}

\begin{table*}
  \centering
  \small
  \setlength{\tabcolsep}{.061in}
\begin{tabular}{l r@{+}l
                r@{ \scriptsize$\pm$ }l r@{ \scriptsize$\pm$ }l l@{ \scriptsize$\pm$ }l
                r@{ \scriptsize$\pm$ }l r@{ \scriptsize$\pm$ }l l@{ \scriptsize$\pm$ }l}
\toprule
  &
  \multicolumn{2}{c}{\#ex.} &
  \multicolumn{6}{c}{Qwen3-4B} &
  \multicolumn{6}{c}{Qwen3-14B} \\ \cmidrule(lr){4-9} \cmidrule(lr){10-15}
& G & A &
  \multicolumn{2}{c}{P} & \multicolumn{2}{c}{R} & \multicolumn{2}{c}{F1} &
  \multicolumn{2}{c}{P} & \multicolumn{2}{c}{R} & \multicolumn{2}{c}{F1} \\ \midrule

Few-shot baseline              & 1 & 0 & 28.7 & {\scriptsize 0.99} & 17.8 & {\scriptsize 1.21} & 22.0$^{\dagger}$ & {\scriptsize 1.19} & 31.6 & {\scriptsize 0.96} & 40.3 & {\scriptsize 1.47} & 35.4$^{\dagger}$ & {\scriptsize 1.02} \\ \midrule


Including \emph{A}dditional examples \\

~~~Paraphrased with LLM
& 1 & 4 & 30.8 & {\scriptsize 1.53} & 08.6 & {\scriptsize 0.96} & 13.5$^{\dagger}$ & {\scriptsize 1.28} & 42.5 & {\scriptsize 1.17} & 24.7 & {\scriptsize 0.73} & 31.2$^{\dagger}$ & {\scriptsize 0.68} \\
& 1 & 9 & 31.4 & {\scriptsize 2.17} & 09.1 & {\scriptsize 0.83} & 14.1$^{\dagger}$ & {\scriptsize 1.16} & 41.0 & {\scriptsize 1.27} & 23.0 & {\scriptsize 0.66} & 29.5$^{\dagger}$ & {\scriptsize 0.82} \\

~~~Generated with LLM
& 1 & 4 & 26.8 & {\scriptsize 0.50} & 24.2 & {\scriptsize 1.34} & 25.4$^{\dagger}$ & {\scriptsize 0.92} & 37.3 & {\scriptsize 1.02} & 40.2 & {\scriptsize 1.73} & {\bf 38.7}$^{*}$ & {\scriptsize {1.11}} \\
& 1 & 9 & 27.8 & {\scriptsize 1.08} & 24.7 & {\scriptsize 1.14} & {\bf 26.1}$^{\dagger}$ & {\scriptsize {0.84}} & 38.6 & {\scriptsize 1.42} & 36.4 & {\scriptsize 1.55} & 37.5$^{\dagger}$ & {\scriptsize 1.10} \\

~~~Retrieved (closest) using \\

~~~~~~\sbert{} representations
& 1 & 4 & 24.5 & {\scriptsize 1.47} & 15.4 & {\scriptsize 1.67} & 18.9$^{\dagger}$ & {\scriptsize 1.56} & 31.1 & {\scriptsize 1.14} & 36.0 & {\scriptsize 1.41} & 33.4$^{\dagger}$ & {\scriptsize 1.05} \\
& 1 & 9 & 24.9 & {\scriptsize 1.10} & 15.9 & {\scriptsize 0.88} & 19.4$^{\dagger}$ & {\scriptsize 0.83} & 30.4 & {\scriptsize 1.09} & 31.1 & {\scriptsize 0.87} & 30.7$^{\dagger}$ & {\scriptsize 0.76} \\

~~~~~~~~~+ hybrid
& 1 & 4 & 25.9 & {\scriptsize 0.58} & 23.4 & {\scriptsize 1.31} & 24.6$^{\dagger}$ & {\scriptsize 0.81} & 39.3 & {\scriptsize 0.70} & 39.4 & {\scriptsize 1.48} & \bf 39.3 & {\scriptsize 0.70} \\
& 1 & 9 & 26.6 & {\scriptsize 0.91} & 25.4 & {\scriptsize 0.79} & {\bf 26.0}$^{\dagger}$ & {\scriptsize 0.57} & 39.3 & {\scriptsize 0.80} & 35.7 & {\scriptsize 1.40} & 37.4$^{\dagger}$ & {\scriptsize 0.86} \\ \addlinespace

~~~~~~Semantic rule representations
& 1 & 4 & 23.9 & {\scriptsize 0.48} & 23.3 & {\scriptsize 0.56} & 23.6$^{\dagger}$ & {\scriptsize 0.43} & 33.3 & {\scriptsize 1.15} & 38.6 & {\scriptsize 1.49} & 35.8$^{\dagger}$ & {\scriptsize 1.21} \\
& 1 & 9 & 22.9 & {\scriptsize 0.56} & 25.1 & {\scriptsize 0.68} & 23.9$^{\dagger}$ & {\scriptsize 0.50} & 33.0 & {\scriptsize 1.66} & 33.9 & {\scriptsize 2.08} & 33.4$^{\dagger}$ & {\scriptsize 1.76} \\

~~~~~~~~~+ hybrid
& 1 & 4 & 26.6 & {\scriptsize 0.87} & 26.7 & {\scriptsize 0.61} & 26.6$^{*}$ & {\scriptsize 0.54} & 39.4 & {\scriptsize 1.04} & 39.4 & {\scriptsize 1.27} & {\bf 39.4} & {\scriptsize 0.89} \\

& 1 & 9 & 26.2 & {\scriptsize 0.60} & 28.9 & {\scriptsize 0.91} & {27.5} & {\scriptsize {0.48}} & 40.0 & {\scriptsize 1.11} & 36.2 & {\scriptsize 1.12} & 38.0$^{\dagger}$ & {\scriptsize 0.82} \\

~~~~~~~~~and $k$-means clustering
& 1 & 4 & 22.8 & {\scriptsize 0.89} & 28.5 & {\scriptsize 1.21} & 25.3$^{\dagger}$ & {\scriptsize 0.96} & 29.5 & {\scriptsize 0.81} & 46.3 & {\scriptsize 1.18} & 36.0$^{\dagger}$ & {\scriptsize 0.83} \\
& 1 & 9 & 22.1 & {\scriptsize 0.75} & 30.6 & {\scriptsize 0.98} & 25.7$^{\dagger}$ & {\scriptsize 0.70} & 29.7 & {\scriptsize 0.89} & 45.1 & {\scriptsize 1.59} & 35.8$^{\dagger}$ & {\scriptsize 0.96} \\

~~~~~~~~~~~~+ hybrid
& 1 & 4 & 25.7 & {\scriptsize 1.05} & 26.6 & {\scriptsize 0.68} & 26.1$^{\dagger}$ & {\scriptsize 0.70} & 37.9 & {\scriptsize 0.94} & 40.5 & {\scriptsize 1.64} & 39.1 & {\scriptsize 1.15} \\
& 1 & 9 & 26.4 & {\scriptsize 0.69} & 28.9 & {\scriptsize 1.88} & {\bf 27.6} & {\scriptsize 1.20} & 38.7 & {\scriptsize 1.19} & 37.0 & {\scriptsize 1.36} & 37.8$^{\dagger}$ & {\scriptsize 0.84} \\ \bottomrule

\end{tabular}

  \caption{Results obtained with \tacred{} using in-context learning in the $1$-shot setting (i.e., one \emph{G}old example).
  Including \emph{A}dditional examples is always beneficial.
  Our hybrid approach to obtain additional examples
  (i.e., coupling LLM-generated examples with our semantic rule-based retrieval)
  yields the best results.
  Only the 4B model benefits from nine instead of four additional examples,
  and the hybrid approach is always beneficial.
  $^{*}$ and $^{\dagger}$ indicate the best method in the whole column is statistically significantly better at $p < 0.05$ and $p < 0.01$, respectively, 
  based on one-sided paired bootstrap test with replacement on F1 scores.
  }  
  \label{t:results_fs-tacred_main}
\end{table*}

Tables \ref{t:results_fs-tacred_main} and \ref{t:results_fs-fewrel_main}
present the main results on \tacred{} and \fewrel{} with \qwensmall{} and \qwenlarge{}.
The appendices present results with additional strategies to obtain additional examples (other clusterings, summarizing, etc.)
on the same LLMs and datasets (Tables \ref{t:results_fs-tacred_full_qwen} and \ref{t:results_fs-fewrel_full_qwen})
as well as \gemmasmall{} and \gemmalarge{} (Tables \ref{t:results_fs-tacred_full_gemma} and \ref{t:results_fs-fewrel_full_gemma}).
We draw several observations from these results.

\paragraph{Structured semantic information helps select better examples for ICL}
In three out of the four scenarios investigated using hybrid systems (i.e., two LLMs evaluated on two datasets)
using examples retrieved with our semantic rule representations yield better overall results 
than all alternatives ($1$-shot baseline, using LLMs to generate or paraphrase, retrieving with SBERT representations).
We conjecture that this is due to the fact that the examples retrieved with our rules
(a)~are realistic, as they are retrieved from a collection of real texts;
(b) preserve the relation faithfulness better due to the underlying rules;
and
(c) are more diverse (see Section \ref{sec:qualitative_analysis}) than LLM-generated and LLM-paraphrased examples,
which tend to stay closer to the original support sentence.
An exception occurs with \qwenlarge{} on \fewrel{}, where the LLM-generated examples slightly outperform our approach.
This suggests that larger models sometimes incorporate more diversity in their generated examples than smaller ones.
However, overall, structured semantic information yields better example selection for ICL in most settings.

\paragraph{Retrieved examples complement LLM-generated examples}
The hybrid strategy 
consistently ranks at or near the top (F1)
with
the smaller models on both datasets
and
one large model (\qwenlarge) on \tacred{}
(Tables \ref{t:results_fs-tacred_main}, \ref{t:results_fs-fewrel_main}, \ref{t:results_fs-tacred_full_gemma} and \ref{t:results_fs-fewrel_full_gemma}).
This suggests that the two sources provide complementary benefits.

Table \ref{t:results_fs-tacred_example_picks} provides insights into the selection process among LLM-generated and retrieved examples.
On average,
one example from the retrieved ones is picked to form the additional four-shot pool; 
and one to three examples from the retrieved ones are picked to form the nine-shot pool.
This suggests the two approaches provide complementary strengths: retrieved examples bring realistic, relation-faithful contexts,
while LLM-generated ones are simpler with strong semantic coverage around the support sentence.
Combining these methods with hybrid strategy results in larger gains than using either source alone,
underlining that hybrid selection is the most reliable way to boost ICL performance for few-shot relation extraction.

\paragraph{Semantic rule representations yield better results than SBERT and paraphrasing with LLMs}
Our semantic rule-based retrieval is consistently stronger than LLM paraphrasing and \sbert{}-based retrieval.
In both \tacred{} and \fewrel{}, paraphrased examples and \sbert-based retrieved examples lead to lower F1s, particularly for smaller models.
On the other hand, semantic rule representation-based retrieved examples maintain better overall F1 (Tables \ref{t:results_fs-tacred_main}, \ref{t:results_fs-fewrel_main}, \ref{t:results_fs-fewrel_full_gemma}, and \ref{t:results_fs-tacred_full_gemma}).
We observe that paraphrasing tends to result in additional examples that remain too close to the original support sentence,
limiting diversity and reducing their effectiveness.
In contrast, \sbert{}-based retrieval often selects examples that are topically similar but do not accurately preserve the target relation.
Overall, semantic rule representations allow us to retrieve more diverse and relation-faithful instances,
leading to better generalization in ICL.

\paragraph{Semantic rule representations work better with smaller LLMs}
The gains from including examples retrieved with our semantic rule representations are clear for the smaller \qwensmall\ and \gemmasmall\ models,
and modest for Qwen3-14B (Tables \ref{t:results_fs-tacred_main}, \ref{t:results_fs-fewrel_main}, \ref{t:results_fs-fewrel_full_gemma}, and \ref{t:results_fs-tacred_full_gemma}).
The larger \gemmalarge{} model shows no improvement with examples retrieved with any retrieval strategy. 
This suggests that structured example selection is most helpful when model capacity is limited,
whereas the larger models exhibit more inconsistent behavior.
Some models benefit more from internal example generation than retrieval,
whereas others fail to benefit from either approach. 

\begin{table*}[h!]
  \centering
  \small
\begin{tabular}{l r@{+}l
                r@{ \scriptsize $\pm$ }l r@{ \scriptsize $\pm$ }l r@{ \scriptsize $\pm$ }l
                r@{ \scriptsize $\pm$ }l r@{ \scriptsize $\pm$ }l r@{ \scriptsize $\pm$ }l}
\toprule
  &
  \multicolumn{2}{c}{\#ex.} &
  \multicolumn{6}{c}{\tacred} \\ \cmidrule(lr){4-9}
& G & A &
  \multicolumn{2}{c}{P} & \multicolumn{2}{c}{R} & \multicolumn{2}{c}{F1} \\ \midrule

MNAV~\cite{fs-tacred}        & 1 & 0 & \multicolumn{2}{c}{-} & \multicolumn{2}{c}{-} & 12.4 & {\scriptsize 1.01}   \\

OdinSynth~\cite{odinsynth}   & 1 & 0 & 23.5 & {\scriptsize 1.46} & 11.5 & {\scriptsize 1.02} & 15.4 & {\scriptsize 1.21} \\
                     
CKPT~\cite{CKPT}             & 1 & 0 & \multicolumn{2}{c}{-} & \multicolumn{2}{c}{-} & 15.1 & {\scriptsize 1.12}  \\

Anchor+gen. rules~\cite{anchorword} & 1 & 0 & 19.6 & {\scriptsize 0.63} & 31.9 & {\scriptsize 1.04} & 24.2 & {\scriptsize 0.72} \\

SoftRules~\cite{softmatcher}  & 1 & 0 & 33.5 & {\scriptsize 1.47} & 19.7 & {\scriptsize 1.14} & 24.8 & {\scriptsize 1.22} \\ \addlinespace
                            
\qwensmall, hybrid (ours) & 1 & 9 & 26.4 & {\scriptsize 0.69} & 28.9 & {\scriptsize 1.88} & \textbf{27.6} & {\scriptsize {1.20}} \\ \bottomrule

\end{tabular}

  \caption{
Comparison of our best approach with Qwen3-4B on \tacred{} (Table \ref{t:results_fs-tacred_main})
and previous work.
All methods only use one \emph{G}old support example.
Even with the smaller (4B) Qwen, in-context learning with the gold example and nine \emph{A}dditional examples retrieved via our hybrid approach yields the new state of the art.
The same approach using \qwenlarge{} yields stronger results (F1: 37.8, Table \ref{t:results_fs-tacred_main}).
}
  \label{t:results_previous_work}
\end{table*}

\paragraph{ICL with additional examples retrieved with semantic rule representations outperforms previous work}
Table \ref{t:results_previous_work} compares previous work and
our results with \qwensmall{} on \tacred{} (hybrid approach, bottom row in Table \ref{t:results_fs-tacred_main}).
These prior methods are dedicated few-shot relation extraction approaches and do not rely on in-context learning;
at inference time, they use one gold shot and no additional examples.
Additionally, their approaches are based on fine-tuning relatively smaller models such as BERT~\citep{bert}, RoBERTa~\citep{roberta}, and CodeT5~\citep{codet5}.

Using one gold shot plus nine retrieved examples in our hybrid approach outperforms all prior work by a substantial margin
(F1: 27.6 vs. 24.8).
The improvements are even larger if we switch the LLM with \qwenlarge{} (F1: 37.8, Table \ref{t:results_fs-tacred_main}).
These results show that
adding genuine examples retrieved from unannotated corpora consistently
outperforms prior work without requiring additional gold examples or model fine-tuning.

\section{Qualitative Analyses}
\label{sec:qualitative_analysis}

To shed more light on the full system and the benefits of retrieving examples
with semantic rule representations,
we carried out multiple qualitative analyses.
We summarize these analyses below. 

\paragraph{Structured semantic retrieval yields more informative examples}
Table \ref{t:diversity_examples} illustrates the additional examples 
obtained with LLMs and retrieved from corpora.
LLM-generated examples are fluent but tend to
be generic,
mention well-known entities,
and closely follow the same wording as the support sentence.
On the other hand, paraphrasing 
mostly repeats the support sentence, offering limited differences beyond swapping a handful of words.
\sbert{}-based retrieved examples are realistic but often only topically similar; additionally, they often miss the relation at hand.
In contrast, due to its semantic underpinning,
the examples retrieved with semantic rule representations
include varied contexts that better express the target relation, making them more useful for ICL.

\paragraph{Diverse examples do not always yield better results}
As shown in Table \ref{t:diversity_comparison},
some methods (e.g., selected strategies under semantic rule-based retrieval) result in very diverse example sets
~(low token \% overlap and cosine similarity),
yet the most diverse examples are not always the top performers (Tables \ref{t:results_fs-tacred_main} and \ref{t:results_fs-fewrel_main}).
Although diversified examples improve over the baseline,
extremely diversified examples can drift away from the relation at hand, which negatively impacts performance.

\paragraph{Additional examples are not always beneficial}
For most settings, adding additional examples improves over the few-shot baseline, but the large model (\gemmalarge) is an exception~(Table~\ref{t:results_fs-tacred_full_gemma} and~\ref{t:results_fs-fewrel_full_gemma}).
Indeed, on both datasets, none of the strategies (LLM-based, retrieval-based, or hybrid) improves over its $1$-shot baseline performance, and several even degrade it.
This indicates that larger models can be sensitive to prompt augmentation:
they may already have enough task understanding that extra examples offer little benefit or can even be detrimental.

\paragraph{Balanced similarity-diversity trade-off works best}
Table \ref{t:diversity_comparison} shows a clear trade-off between staying close to the (gold) support example and maintaining diversity among additional examples.
LLM paraphrases are very similar to the gold support, having high token overlap and cosine similarity, which indicate low diversity.
On the other hand, the retrieved example sets using $k$-means and its variants maximize diversity but also can drift away from the support sentence and the relation.
The best performance comes from the middle ground, i.e., the hybrid selection, which retains enough similarity to preserve the relation (through the generated examples) while introducing diversity (through the more realistic retrieved examples). 

\paragraph{Identifying multiple relations at once with a single ICL hurts performance}
Table \ref{t:results_all_relation_single_prompt} shows that using a single prompt for multi-class relation classification yields much lower F1 than
prompting each relation separately with a binary decision prompt (i.e., does this relation hold or not?).
In the multi-class setting, especially for the smaller models, we observe errors such as predicting misspelled relation labels or hallucinated relations.
We therefore use five binary prompts in our main experiments and did not pursue the single-prompt option further.

\paragraph{NER-filtering as preprocessing consistently improves results}
Table~\ref{t:results_without_ner_filter}
presents results with and without the NER filtering step.
Across datasets and models, the NER filtering consistently improves the F1 scores.
Furthermore, removing this preprocessing step substantially increases computational cost, as more relations must be prompted per episode.
Overall, NER filtering improves precision without a noticeable drop in recall.

\paragraph{Computational Efficiency and Runtime Analysis}
Our best approach, the hybrid method, increases the wall-clock runtime by only about 0.5\% per episode compared to the LLM-generated baseline. 
In one episode, retrieval and clustering are performed for five relation instances, resulting in a total cost of 6.9s (1.38s × 5); 
this cost is independent of the underlying LLM model size, as it relies solely on embedding-based operations. 
Similarly, LLM-based example generation using Qwen3-4B is invoked five times, for a total of 8.95s (1.79s × 5). 
In contrast, the additional example selection step used in the hybrid approach is applied only once per episode, 
generates only a small list of 4 or 9 IDs, and adds just 0.049s of overhead, 
leading to negligible computational cost overall.
\section{Conclusions}
\label{sec:conclusion}




$1$-shot relation extraction is a challenging problem that is however critical in domains where response time matters, e.g., intelligence or medical.
We study how to improve in-context learning in this setting by selecting additional examples 
that effectively transform a $1$-shot setup into a $5$- or $10$-shot setting without supervision or model training.
Specifically, we propose an example selection framework that retrieves relation-faithful examples using 
structured syntactic-semantic representations, promotes diversity through clustering, and complements retrieval with LLM-generated examples.
Together, these strategies yield a robust hybrid selection method that balances similarity to the gold example 
with diversity across examples.

Our hybrid example selection strategy transfers well across benchmarks 
(\tacred{} and \fewrel{}) and LLMs (\qwensmall{}, \qwenlarge{}, \gemmasmall{}).
Overall, our analyses show that explicitly balancing (a)~similarity to the gold example via LLM-generated examples 
and (b)~diversity via syntactic-semantic retrieval is key to effective example selection for in-context learning.

\section*{Limitations}
\label{sec:limitations}

While our proposed approach is effective, some practical limitations remain. 
Not all relations naturally align across corpora; therefore, our experiments focus on the subset of FewRel relations that align with TACRED. 
In addition, different LLMs can respond differently to the same example set, suggesting that example selection could be further tailored to individual models for optimal performance. 
Finally, although semantic rule representations provide a principled way to retrieve examples, occasional inaccuracies in these rules may introduce less relevant demonstrations. 
Addressing these aspects offers directions for future work.
\section*{Acknowledgements}
\label{sec:acknowledgement}

We thank Robert Vacareanu for sharing data resources and for providing reference code and documentation that clarified the processing of the UMBC Index.

Mihai Surdeanu declares a financial interest in lum.ai. 
This interest has been properly disclosed to the University of Arizona Institutional Review Committee and is
managed in accordance with its conflict of interest policies.

\nocite{*}
\bibliography{custom}

\appendix
\section{Appendix}
\label{sec:appendix}



\subsection{Hyperparameter Tuning and Retrieval Efficiency}
\label{subsec:hyperparameters-retrieval}
We use the \tacred\ development episodes as held-out data to tune hyperparameters.
In practice, the only tunable parameter in our framework is the similarity threshold $\tau$, which controls the pool of candidate examples selected for clustering.
We tune $\tau$ on \tacred\ development episodes and fix it to $\tau = 0.6$ in all experiments.
The same threshold value is applied when evaluating on \fewrel\ , and no additional hyperparameters are tuned on the \fewrel\ development set.
All other components, including the clustering algorithms and cluster selection strategies, use fixed settings as described in Section~\ref{sec:example-selection-fsre}.

To enable efficient retrieval at scale, cosine similarity search over precomputed sentence and rule embeddings is implemented using FAISS~\cite{faiss}, allowing all candidate similarities to be computed offline and reused during inference.
This design keeps inference efficient and avoids introducing additional tunable parameters.

\subsection{Decoding and Inference Details}
\label{subsec:decoding_details}
The relation inference and named entity recognition tasks are formulated as binary classifications 
using the logits of the first generated token, restricted to \texttt{``yes''} and \texttt{``no''}.
A prediction of \texttt{``yes''} is made when $P(\texttt{yes}) \geq\ P(\texttt{no})$, and \texttt{``no''} otherwise.
For all generation-related components, we use the default decoding configurations of the underlying language models.
For Qwen models, we use \texttt{do\_sample}=\texttt{True}, \texttt{temperature}=$0.6$, \texttt{top\_p}=$0.95$, and \texttt{top\_k}=$20$, 
following the model defaults.
For Gemma models, we use \texttt{do\_sample}=\texttt{True}, \texttt{top\_p}=$0.95$, \texttt{top\_k}=$64$, and the default temperature setting~(=$1.0$).
We do not truncate input prompts, while generated outputs are capped at a maximum of 1,000 tokens, 
which is sufficient for all prompts and tasks considered in this work.
These settings follow standard model usage and avoid introducing additional decoding hyperparameters that could confound comparisons across strategies.



\begin{figure*}
\centering
\small
\begin{tabularx}{\textwidth}{X}
\toprule
\texttt{You are given below
a Relation name,
a Description of the relation between brackets,
N Support sentences exemplifying the relation,
and a Query sentence.} \\\ \\

\texttt{A relation connects the Subject and the Object entities.
The Subject and the Object entities are indicated with the subject and object tags respectively.
You need to decide whether the relation holds between the Subject and the Object entities in the Query sentence.} \\\ \\

\texttt{Relation name: "\#RELATION\#" (\#RELATION\_DESCRIPTION\#)} \\ 
\texttt{Support Sentence 1: \#SUPPORT\_SENTENCE\_1\#} \\ 
\texttt{Support Sentence 2: \#SUPPORT\_SENTENCE\_2\#} \\ 
\texttt{\ldots} \\ 
\texttt{Support Sentence N: \#SUPPORT\_SENTENCE\_N\#} \\\ \\

\texttt{Query Sentence: \#QUERY\_SENTENCE\#} \\\ \\

\texttt{If the relation holds between the Subject and Object entities in the Query sentence, say ``yes,'' otherwise say ``no.'' Just output ``yes'' or ``no,'' and nothing else.} \\
\bottomrule
\end{tabularx}
\caption{
Prompt template used for relation extraction using the binary approach ($N \in \{1, 5, 10\}$ in our experiments).
Given 
(a) a relation name and description,
(b) a set of support sentences that exemplify the relation,
and
(c) a query sentence, 
the model is asked to decide whether the relation holds between the subject and object entities in the query sentence.
}
\label{t:prompt_one-relation-inference}
\end{figure*}

\begin{figure*}
\small
\centering
\begin{tabularx}{\textwidth}{X}
\toprule
\texttt{You are given below
a sentence,
an entity contained within the sentence,
and an entity type:}\\\ \\

\texttt{Sentence: \#SENTENCE\#} \\
\texttt{Entity: \#ENTITY\#} \\
\texttt{Entity Type: \#ENTITY\_TYPE\#} \\\ \\

\texttt{Your task is to decide whether the Entity in the context of the Sentence either:}\\
\texttt{1. belongs to the entity type "\#ENTITY\_TYPE\#"}\\
\texttt{or,}\\
\texttt{2. is a mention that points to an entity that belongs to the entity type "\#ENTITY\_TYPE\#" (such as a pronoun or other co-referring expression)}\\\ \\

\texttt{Only answer ``yes'' or ``no,'' nothing else.}\\
\bottomrule
\end{tabularx}
\caption{
Prompt template used for validating compatibility between
the (expected) named entity types of a relation type
and the entities in a query sentence (Section \ref{sec:few-shot-relation-extraction-task}).
The entity types considered in this work are PERSON, LOCATION, ORGANIZATION, DATE, CITY, COUNTRY, STATE, and PROVINCE).
}
\label{t:prompt-entity-resolution}
\end{figure*}

\begin{figure*}
\centering
\small
\begin{tabularx}{\textwidth}{X}
\toprule
\texttt{You are given below
five Relation names,
the Description of the relations between brackets,
N Support sentences exemplifying each relation,
and a Query sentence.} \\\ \\

\texttt{A relation connects the Subject and the Object entities.
The Subject and the Object entities are indicated with the subject and object tags respectively.
You need to decide whether the relation holds between the Subject and the Object entities in the Query sentence.} \\\ \\

\texttt{Relation name: "\#RELATION1\#" (\#RELATION\_DESCRIPTION\#)} \\ 
\texttt{Support Sentence 1: \#SUPPORT\_SENTENCE\_1\#} \\ 
\texttt{Support Sentence 2: \#SUPPORT\_SENTENCE\_2\#} \\ 
\texttt{\ldots} \\ 
\texttt{Support Sentence N: \#SUPPORT\_SENTENCE\_N\#} \\\ \\

\texttt{Relation name: "\#RELATION2\#" (\#RELATION\_DESCRIPTION\#)} \\ 
\texttt{Support Sentence 1: \#SUPPORT\_SENTENCE\_1\#} \\ 
\texttt{Support Sentence 2: \#SUPPORT\_SENTENCE\_2\#} \\ 
\texttt{\ldots} \\ 
\texttt{Support Sentence N: \#SUPPORT\_SENTENCE\_N\#} \\\ \\

\texttt{\ldots}\\\ \\

\texttt{Relation name: "\#RELATION5\#" (\#RELATION\_DESCRIPTION\#)} \\ 
\texttt{Support Sentence 1: \#SUPPORT\_SENTENCE\_1\#} \\ 
\texttt{Support Sentence 2: \#SUPPORT\_SENTENCE\_2\#} \\ 
\texttt{\ldots} \\ 
\texttt{Support Sentence N: \#SUPPORT\_SENTENCE\_N\#} \\\ \\

\texttt{If the relation holds between the Subject and Object entities in the Query sentence, say ``yes,'' otherwise say ``no.'' Just output ``yes'' or ``no,'' and nothing else.} \\
\bottomrule
\end{tabularx}
\caption{
Prompt template used for relation extraction with the multi-class approach, which we evaluate only for the $1$-shot baseline ($N=1$), with five candidate relations per episode.
Given 
(a) five relation names and their descriptions,
(b) a set of support sentences that exemplify each relation,
and
(c) a query sentence, 
the model is asked to decide which relation, if any, holds between the subject and object entities in the query.
}
\label{t:prompt_multi-relation-inference}
\end{figure*}

\begin{table*}
  \centering
  \small 
  \begin{tabularx}{\textwidth}{p{.8in} X}
\toprule
 
Method & Additional Examples \\ \midrule
 
Paraphrased with LLM
& 1. \sbj{New Fabris} company director \obj{Pierre Reau} stated that the company can currently only offer between 10,000 and 15,000 euros in redundancy to workers with 20 or more years of experience. \\
& 2. \sbj{New Fabris} director \obj{Pierre Reau} mentioned that the firm is currently only able to pay between 10,000 and 15,000 euros in redundancy to employees with 20 or more years of experience. \\
& 3. \sbj{New Fabris} company director \obj{Pierre Reau} explained that the company can only currently pay between 10,000 and 15,000 euros in redundancy to workers with 20 or more years of service. \\
& 4. \sbj{New Fabris} director \obj{Pierre Reau} said that the company is currently only able to provide between 10,000 and 15,000 euros in redundancy to employees with 20 or more years of experience. \\ \midrule

Generated with LLM
& 1. \sbj{GlobalTech} CEO \obj{Emma Lee} highlighted that the company's top employees are driving innovation and growth. \\
& 2. \sbj{GreenEnergy} executive \obj{James Carter} mentioned that the firm's top members are leading sustainable development initiatives. \\
& 3. \sbj{HealthNova} director \obj{Dr. Laura Martinez} stated that the organization's top employees are focused on patient care and research. \\
& 4. \sbj{ArtisanWorks} founder \obj{Robert Kim} noted that the company's top members are responsible for creative direction and product design. \\ \midrule

Retrieved \sbert
& 1. According to \sbj{Microtune}, many of the laid off employees have been hired by \obj{TFS}, which has agreed to address and build demand for radio frequency subsystems. \\
& 2. \obj{Derek Twigg} : In line with the Government's policy on controlling public sector pay, the Royal Fleet Auxiliary ( RFA ) pay submission has been subject to close scrutiny by the \sbj{Ministry of Defence and HM Treasury}. \\
& 3. The new Board will work with the \sbj{FSA} to merge six existing compensation schemes into a single one stop shop, as proposed under the Financial Services and Markets \obj{Bill}, currently before Parliament. \\
& 4. "We are expecting the worst after August 14, when there will be retrenchment and the golden handshake will be offered to bank employees ahead of privatization," said \obj{Habibuddin Junaidi}, president of \sbj{Habib Bank General Workers Front} Pakistan. \\\midrule

Retrieved with semantic rule representations
& 1. Among the representative works of the Bai people are Transit Star Catalogue for Time Determination by the \sbj{Ming Dynasty} scholar \obj{Zhou Silian}, Collection of Secret Prescriptions by Chen Dongtian and Tested Prescriptions by Li Xingwei.\\
& 2. \sbj{Sony Pictures} Classics co-presidents \obj{Michael Barker} and Tom Bernard dropped in on a small dinner at local hotspot Prego, celebrating Alice Wu's "Saving Face" last week, and quickly closed a deal for the Discovery section film.\\
& 3. "Options under consideration include regular and expanded sampling for hazardous substances such as anthrax, opening packages for inspection, and high-tech alternatives, including digitizing the mail," \sbj{House} Administration Chairman \obj{Bob Ney} ( R-Ohio ) wrote in a recent letter to his colleagues.\\
& 4. \sbj{Elon University} freshman \obj{Elizabeth Easterly} presented research on body dissatisfaction, self-esteem and exercise motivators among college-age women.\\ \midrule

\multirow{2}{.8in}{Retrieved with semantic rule representations and $k$-means clustering}
& 1. For further information, contact \obj{Mike Smart}, the \sbj{Bureau}'s Senior Policy Advisor ( you can address him in German if you wish, as well as in English, French, and Spanish, and what not else ). \\
& 2. "IPv6 is an emerging technology that will increase in significance as the demand for IP addresses grows," says \obj{Lawrence Orans}, senior analyst, \sbj{Gartner, Inc.} "We anticipate that this demand will be strongest in the Asia-Pacific region, which owns far fewer registered IP addresses than does North America." \\
& 3. After graduating from Strathclyde, I obtained a PhD at the University of Glasgow and continued as a researcher in the lab of Dr \obj{Jim Brewer} ( himself a \sbj{Strathclyde Immunology} graduate ) looking at ways to improve the efficacy of vaccines. \\
& 4. \obj{Lieberman}, a three-term senator and the \sbj{Democratic} vice presidential candidate in 2000, lost by a 52 to 48 percent margin to Ned Lamont, an heir to the Lamont family fortune and multi-millionaire businessman, who ran as an opponent of the war in Iraq. \\
\bottomrule
 
\end{tabularx}

  \caption{
  Examples of additional support examples obtained with different methods (Section \ref{sec:example-selection-fsre}) given the following original support sentence:
  \emph{\sbj{New Fabris} company director \obj{Pierre Reau} said that currently the firm would only able to pay 
between 10,000 and 15,000 euros in redundancy to workers with 20 or more years experience , and some junior workes would get just 3,000 euros} (relation: org:top\_members/employees).
  Additional examples obtained with the LLM (generated from scratch or by paraphrasing) tend to be near duplicates and semantically nearly identical to the original support example.
  On the other hand, retrieval-based methods provide additional examples with more diverse contexts.
  In particular, examples retrieved using semantic rule representations and $k$-means clustering show greater lexical and structural diversity while preserving the target relation.  
}
  \label{t:diversity_examples}
\end{table*}

\begin{figure*}
\centering
\small
\begin{tabularx}{\textwidth}{X}
\toprule
\texttt{You are given below a Relation name, a Description of the relation, and a support sentence exemplifying the relation.}\\\ \\

\texttt{A relation connects two entities: the Subject and the Object entities in the sentence. }
\texttt{The Subject and the Object entities are indicated with the <subject>..</subject> and <object>..</object> tags respectively.} 
\\\ \\
\texttt{Relation name: "\#RELATION\#"} \\
\texttt{Relation description: "\#RELATION\_DESCRIPTION\#"} \\
\texttt{Support Sentence: \#SUPPORT\_SENTENCE\#} \\
\\
\texttt{Your task is to generate N paraphrases of the support sentence that hold the same relation between the same Subject and Object entities.
In each paraphrase, you must include the subject and object tags to identify the Subject and the Object entities.} \\
\\
\texttt{Output in the following format:} \\

\texttt{01: your 1st paraphrased sentence} \\
\texttt{02: your 2nd paraphrased sentence} \\
\ldots... \\
\texttt{N: your Nth paraphrased sentence} \\
\bottomrule
\end{tabularx}
\caption{
Prompt template used for obtaining additional examples by paraphrasing with an LLM the original (gold) support sentence.
Given a relation name, its description, and a support sentence, the model is instructed to generate multiple paraphrases that preserve the relation.
}
\label{t:prompt-paraphrase}
\end{figure*}

\begin{figure*}
\small
\centering
\begin{tabularx}{\textwidth}{X}
\toprule
\texttt{You are given below
a Relation name,
the Description of the relation,
and a support sentence exemplifying the relation.}\\\ \\

\texttt{A relation connects two entities: the Subject and the Object entities in the sentence. }
\texttt{The Subject and the Object entities are indicated with the <subject>..</subject> and <object>..</object> tags respectively.} 
\\\ \\
\texttt{Relation name: "\#RELATION\#"} \\
\texttt{Relation description: "\#RELATION\_DESCRIPTION\#"} \\
\texttt{Support Sentence: \#SUPPORT\_SENTENCE\#} \\
\\
\texttt{Your task is to generate N completely different new examples that hold the same relation. You must follow these guidelines:}

\texttt{1. In each example, include subject and object tags to identify the Subject and the Object entities.} \\
\texttt{2. To increase diversity, use different words, phrases, and sentence structures across different examples.} \\
\\
\texttt{Output in the following format:} \\

\texttt{01: your 1st example sentence} \\
\texttt{02: your 2nd example sentence} \\
\texttt{...} \\
\texttt{N: your Nth example sentence} \\
\bottomrule
\end{tabularx}
\caption{
Prompt template used for obtaining additional examples by generating them with an LLM.
Given a relation name, its description, and a support sentence, the model is instructed to generate diverse examples that hold the relation.
}
\label{t:prompt-new-example-generation}
\end{figure*}

\begin{figure*}
\small
\centering
\begin{tabularx}{\textwidth}{X}
\toprule
\texttt{You are given below
a Relation name,
a Description of the relation between brackets,
and N Support sentences exemplifying the relation.} \\\ \\

\texttt{A relation connects the Subject and the Object entities. The Subject and the Object entities are marked within the subject and object tags respectively.}\\\ \\

\texttt{Relation name: "\#RELATION\#" (\#RELATION\_DESCRIPTION\#)} \\
\texttt{Support Sentence 1: \#SUPPORT\_SENTENCE\_1\#} \\ 
\texttt{Support Sentence 2: \#SUPPORT\_SENTENCE\_2\#} \\ 
\texttt{\ldots}\\
\texttt{Support Sentence N: \#SUPPORT\_SENTENCE\_N\#} \\\ \\

\texttt{Your task is to pick N/2 support sentences that maximize diversity.
In other words, you should pick support sentences that use different words, phrases, and sentence structures.} \\\ \\

\texttt{Output your best picks as a Python-style list of the N/2 IDs of the support sentences (e.g., [1, 4, 6, 7]).} \\

\texttt{Only output the list and nothing else.} \\
\bottomrule
\end{tabularx}
\caption{
Prompt template used in our hybrid approach to 
select additional examples from the ones
(a)~generated with an LLM
and
(b) retrieved with our semantic rule representations and $k$-means clustering.
In our experiments,
(a) $N=8$ or $N=18$
and
(b) we shuffle the $N$ support sentences in the prompt to avoid positional biases.
}
\label{t:prompt-pick_examples}
\end{figure*}

\begin{figure*}
\centering
\small
\begin{tabularx}{\textwidth}{X}
\toprule
\texttt{You are given a context sentence containing a Subject and an Object entity.} \\\\

\texttt{The Subject and Object entities are marked using <subject> and <object> tags, respectively.} \\\\

\texttt{Your task is to summarize the relation expressed between the Subject and the Object entities in the context.} \\\\

\texttt{Context: \#SUPPORT\_SENTENCE\#} \\\\

\texttt{You must retain the <subject> and <object> tags in the summarized output.} \\\\

\texttt{Only output the summarized relation between the Subject and the Object entities, and nothing else.} \\
\bottomrule
\end{tabularx}

\caption{
Prompt template used for summarizing support sentences.
Given a support sentence exemplifying a relation, 
the model is instructed to produce a compact natural language description of the relation between the subject and object entities,
while retaining the subject and object tags.
}
\label{t:prompt_summariztation}
\end{figure*}

\begin{table*}
  \centering
  \small
\begin{tabular}{l r@{+}l
                r@{ \scriptsize $\pm$ }l r@{ \scriptsize $\pm$ }l r@{ \scriptsize $\pm$ }l
                r@{ \scriptsize $\pm$ }l r@{ \scriptsize $\pm$ }l r@{ \scriptsize $\pm$ }l}
\toprule 
  \multirow{2}{*}{Dataset} &
  \multicolumn{2}{c}{\#ex.} &
  \multicolumn{6}{c}{\qwensmall} &
  \multicolumn{6}{c}{\gemmasmall} \\ \cmidrule(lr){4-9} \cmidrule(lr){10-15}
& G & A &
  \multicolumn{2}{c}{P} & \multicolumn{2}{c}{R} & \multicolumn{2}{c}{F1} &
  \multicolumn{2}{c}{P} & \multicolumn{2}{c}{R} & \multicolumn{2}{c}{F1}\\ \midrule

\tacred & 1 & 0 & 14.6 & {\scriptsize 1.05} & 8.4 & {\scriptsize 0.90} & 10.6 & {\scriptsize 1.00} & 4.4 & {\scriptsize 0.32} & 4.3 & {\scriptsize 0.29} & 4.4 & {\scriptsize 0.30} \\
\fewrel & 1 & 0 & 52.0 & {\scriptsize 2.31} & 16.5 & {\scriptsize 1.36} & 25.0 & {\scriptsize 1.79} & 49.3 & {\scriptsize 7.91} & 1.8 & {\scriptsize 0.18} & 3.4 & {\scriptsize 0.34} \\ 
FS-NYT29 & 1 & 0 & 41.9 & {\scriptsize 0.31} & 64.1 & {\scriptsize 0.69} & 50.7 & {\scriptsize 0.41} & 56.3 & {\scriptsize 1.37} & 21.1 & {\scriptsize 0.82} & 30.7 & {\scriptsize 1.07} \\

\bottomrule
\end{tabular}

  \caption{
  Results obtained with a $1$-shot prompt only providing the subject and object entities and nothing else from the query sentence (Figure~\ref{t:prompt_sbj_obj}).
  This experiment was designed to test the parametric knowledge of the LLMs with respect to the test instances in \tacred{}, \fewrel{}, and FS-NYT29.
  The results are relatively low with \tacred{} and \fewrel{}
  but surprisingly high with FS-NYT29.
  We conclude that the query sentences in FS-NYT29 contain many instances for which parametric knowledge---even in relatively small LLMs---is enough to make the correct inference.
  Thus, we do not consider FS-NYT29 in our main experiments.
}
  \label{t:parametric-test}
\end{table*}

\begin{figure*}
\centering
\small
\begin{tabularx}{\textwidth}{X}
\toprule
\texttt{You are given below a Relation name, a Description of the relation in brackets, a Support sentence exemplifying the relation between the Subject and the Object entities, and a Query (a subject and an object entity).} \\[0.5em]

\texttt{A relation connects the Subject and the Object entities. The Subject and the Object entities are given within the subject and object tags respectively. You need to decide whether the relation holds between the Subject and the Object entities in the given Query.} \\[0.5em]

\texttt{Relation name: ``\#RELATION\#" (\#RELATION\_DESCRIPTION\#)} \\ 
\texttt{Support Sentence 1: \#SUPPORT\_SENTENCE\_1\#} \\ 
\texttt{Support Sentence 2: \#SUPPORT\_SENTENCE\_2\#} \\ 
\texttt{...} \\ 
\texttt{Support Sentence N: \#SUPPORT\_SENTENCE\_N\#} \\[0.5em]

\texttt{Query Subject: \#SUBJECT\#} \\[0.5em]
\texttt{Query Object: \#OBJECT\#} \\[0.5em]

\texttt{If the relation between the Subject and the Object entities in the Query matches the given Relation say ``yes,'' otherwise ``no.''} \\
\bottomrule
\end{tabularx}
\caption{
Prompt template for $1$-shot relation extraction only providing the subject and object entities and nothing
else from the query sentence.
This prompt was designed to test the parametric knowledge of the
LLMs with respect to the test instances in FS-TACRED, FS-FewRel, and FS-NYT29.
}
\label{t:prompt_sbj_obj}
\end{figure*}

\begin{table*}[]
  \centering
  \small
  \setlength{\tabcolsep}{.061in}
\begin{tabular}{l r@{+}l
                r@{ \scriptsize$\pm$ }l r@{ \scriptsize$\pm$ }l l@{ \scriptsize$\pm$ }l
                r@{ \scriptsize$\pm$ }l r@{ \scriptsize$\pm$ }l l@{ \scriptsize$\pm$ }l}
\toprule
  &
  \multicolumn{2}{c}{\#ex.} &
  \multicolumn{6}{c}{\qwensmall} &
  \multicolumn{6}{c}{\qwenlarge} \\ \cmidrule(lr){4-9} \cmidrule(lr){10-15}
& G & A &
  \multicolumn{2}{c}{P} & \multicolumn{2}{c}{R} & \multicolumn{2}{c}{F1} &
  \multicolumn{2}{c}{P} & \multicolumn{2}{c}{R} & \multicolumn{2}{c}{F1} \\ \midrule

Few-shot baseline               & 1 & 0 & 41.7 & {\scriptsize 1.37} & 26.4 & {\scriptsize 1.17} & 32.4$^{\dagger}$ & {\scriptsize 1.14} & 43.5 & {\scriptsize 0.81} & 42.3 & {\scriptsize 0.73} & 42.9$^{\dagger}$ & {\scriptsize 0.52} \\ \midrule
Including \emph{A}dditional examples \\

~~~Paraphrased with LLM               
& 1 & 4 & 44.3 & {\scriptsize 1.91} & 13.4 & {\scriptsize 0.78} & 20.5$^{\dagger}$ & {\scriptsize 1.05} & 42.6 & {\scriptsize 1.17} & 48.8 & {\scriptsize 0.75} & \bf 45.5 & {\scriptsize 0.75} \\
& 1 & 9 & 40.2 & {\scriptsize 2.37} & 13.3 & {\scriptsize 0.97} & 20.0$^{\dagger}$ & {\scriptsize 1.36} & 45.6 & {\scriptsize 0.72} & 43.9 & {\scriptsize 0.51} & 44.7$^{\dagger}$ & {\scriptsize 0.09} \\

~~~Generated with LLM                 
& 1 & 4 & 43.8 & {\scriptsize 0.95} & 29.4 & {\scriptsize 0.75} & {\bf 35.2}$^{\dagger}$ & {\scriptsize 0.75} & 42.7 & {\scriptsize 1.20} & 48.8 & {\scriptsize 0.76} & \bf 45.5 & {\scriptsize 0.78} \\
& 1 & 9 & 42.0 & {\scriptsize 0.96} & 28.4 & {\scriptsize 0.68} & 33.9$^{\dagger}$ & {\scriptsize 0.79} & 45.8 & {\scriptsize 0.92} & 44.2 & {\scriptsize 0.37} & 45.0$^{*}$ & {\scriptsize 0.31} \\

~~~Retrieved (closest) using \\

~~~~~~\sbert{} representations
& 1 & 4 & 43.3 & {\scriptsize 0.94} & 22.4 & {\scriptsize 0.48} & 29.5$^{\dagger}$ & {\scriptsize 0.57} & 43.7 & {\scriptsize 0.95} & 43.0 & {\scriptsize 1.86} & 43.4$^{\dagger}$ & {\scriptsize 1.13} \\
& 1 & 9 & 42.4 & {\scriptsize 1.10} & 21.2 & {\scriptsize 0.45} & 28.2$^{\dagger}$ & {\scriptsize 0.47} & 44.1 & {\scriptsize 1.04} & 39.8 & {\scriptsize 1.52} & 41.8$^{\dagger}$ & {\scriptsize 1.08} \\

~~~~~~~~~+ hybrid
& 1 & 4 & 44.1 & {\scriptsize 1.19} & 29.4 & {\scriptsize 0.68} & {\bf 35.3}$^{\dagger}$ & {\scriptsize 0.86} & 44.0 & {\scriptsize 0.93} & 43.1 & {\scriptsize 1.77} & {\bf 43.5}$^{\dagger}$ & {\scriptsize 1.13} \\
& 1 & 9 & 43.0 & {\scriptsize 0.73} & 29.2 & {\scriptsize 0.51} & 34.7$^{\dagger}$ & {\scriptsize 0.50} & 44.7 & {\scriptsize 1.18} & 39.9 & {\scriptsize 1.53} & 42.1$^{\dagger}$ & {\scriptsize 1.00} \\ \addlinespace

~~~~~~Semantic rule representations
& 1 & 4 & 40.6 & {\scriptsize 1.44} & 33.0 & {\scriptsize 1.30} & 36.4 & {\scriptsize 1.37} & 43.3 & {\scriptsize 0.66} & 45.6 & {\scriptsize 1.63} & \bf 44.4$^{\dagger}$ & {\scriptsize 0.98} \\
& 1 & 9 & 38.1 & {\scriptsize 1.12} & 33.0 & {\scriptsize 0.57} & 35.3$^{\dagger}$ & {\scriptsize 0.78} & 42.7 & {\scriptsize 0.66} & 42.5 & {\scriptsize 1.63} & 42.6$^{\dagger}$ & {\scriptsize 0.98} \\

~~~~~~~~~+ hybrid
& 1 & 4 & 44.1 & {\scriptsize 1.50} & 31.6 & {\scriptsize 1.20} & {\bf 36.8} & {\scriptsize 1.25} & 43.2 & {\scriptsize 0.61} & 45.4 & {\scriptsize 1.49} & 44.2$^{\dagger}$ & {\scriptsize 0.86} \\
& 1 & 9 & 41.9 & {\scriptsize 1.36} & 31.6 & {\scriptsize 0.90} & 36.0 & {\scriptsize 1.06} & 42.8 & {\scriptsize 1.22} & 43.0 & {\scriptsize 1.72} & 42.9$^{\dagger}$ & {\scriptsize 1.19} \\

~~~~~~~~~and $k$-means clustering
& 1 & 4 & 39.1 & {\scriptsize 1.66} & 33.0 & {\scriptsize 0.98} & 35.8$^{*}$ & {\scriptsize 1.22} & 40.3 & {\scriptsize 0.97} & 47.1 & {\scriptsize 1.71} & 43.4$^{\dagger}$ & {\scriptsize 1.03} \\
& 1 & 9 & 35.6 & {\scriptsize 1.23} & 32.4 & {\scriptsize 0.58} & 33.9$^{\dagger}$ & {\scriptsize 0.85} & 39.9 & {\scriptsize 0.99} & 46.3 & {\scriptsize 1.27} & 42.9$^{\dagger}$ & {\scriptsize 0.88} \\

~~~~~~~~~~~~+ hybrid
& 1 & 4 & 43.9 & {\scriptsize 1.49} & 31.6 & {\scriptsize 0.99} & 36.7 & {\scriptsize 1.09} & 40.3 & {\scriptsize 1.14} & 46.7 & {\scriptsize 1.86} & 43.2$^{\dagger}$ & {\scriptsize 1.21} \\
& 1 & 9 & 43.0 & {\scriptsize 1.46} & 32.2 & {\scriptsize 1.29} & {\bf 36.8} & {\scriptsize 1.35} & 39.9 & {\scriptsize 0.82} & 45.9 & {\scriptsize 0.81} & 42.7$^{\dagger}$ & {\scriptsize 0.64} \\ \bottomrule

\end{tabular}

  \caption{Results obtained with \fewrel{} using in-context learning in the $1$-shot setting (i.e., one \emph{G}old example).
  Our hybrid approach to obtain additional examples
  (i.e., coupling LLM-generated examples with our semantic rule-based retrieval)
  yields the best results with \qwensmall{} (F1: +1.6 compared to generating additional examples with an LLM),
  but we only observe benefits on Recall with \qwenlarge{}.
$^{*}$ and $^{\dagger}$ indicate the best method in the whole column is statistically significantly better at $p < 0.05$ and $p < 0.01$, respectively, 
  based on one-sided paired bootstrap test with replacement on F1 scores.}
  \label{t:results_fs-fewrel_main}
\end{table*}

\begin{table*}
  \centering
  \small
  \setlength{\tabcolsep}{.061in}
\setlength{\tabcolsep}{.061in}
\setlength{\tabcolsep}{0.055in}
\begin{tabular}{l r@{+}l
                r@{ \scriptsize$\pm$ }l r@{ \scriptsize$\pm$ }l l@{ \scriptsize$\pm$ }l
                r@{ \scriptsize$\pm$ }l r@{ \scriptsize$\pm$ }l l@{ \scriptsize$\pm$ }l}
\toprule
  &
  \multicolumn{2}{c}{\#ex.} &
  \multicolumn{6}{c}{Qwen3-4B} &
  \multicolumn{6}{c}{Qwen3-14B} \\ \cmidrule(lr){4-9} \cmidrule(lr){10-15} 
& G & A &
  \multicolumn{2}{c}{P} & \multicolumn{2}{c}{R} & \multicolumn{2}{c}{F1} &
  \multicolumn{2}{c}{P} & \multicolumn{2}{c}{R} & \multicolumn{2}{c}{F1} \\ \midrule

Few-shot baseline                 & 1 & 0 & 28.7 & {\scriptsize 0.99} & 17.8 & {\scriptsize 1.21} & 22.0$^{\dagger}$ & {\scriptsize 1.19} & 31.6 & {\scriptsize 0.96} & 40.3 & {\scriptsize 1.47} & 35.4$^{\dagger}$ & {\scriptsize 1.02} \\ \midrule
Including \emph{A}dditional examples \\

~~~Paraphrased with LLM  
& 1 & 4 & 30.8 & {\scriptsize 1.53} & 08.6 & {\scriptsize 0.96} & 13.5$^{\dagger}$ & {\scriptsize 1.28} & {42.5} & {\scriptsize {1.17}} & 24.7 & {\scriptsize 0.73} & 31.2$^{\dagger}$ & {\scriptsize 0.68} \\
& 1 & 9 & {31.4} & {\scriptsize {2.17}} & 09.1 & {\scriptsize 0.83} & 14.1$^{\dagger}$ & {\scriptsize 1.16} & 41.0 & {\scriptsize 1.27} & 23.0 & {\scriptsize 0.66} & 29.5$^{\dagger}$ & {\scriptsize 0.82} \\ 

~~~Generated with LLM 
& 1 & 4 & 26.8 & {\scriptsize 0.50} & 24.2 & {\scriptsize 1.34} & 25.4$^{\dagger}$ & {\scriptsize 0.92} & 37.3 & {\scriptsize 1.02} & 40.2 & {\scriptsize 1.73} & \textbf{38.7}$^{*}$ & {\scriptsize 1.11} \\
& 1 & 9 & 27.8 & {\scriptsize 1.08} & 24.7 & {\scriptsize 1.14} & \textbf{26.1}$^{\dagger}$ & {\scriptsize 0.84} & 38.6 & {\scriptsize 1.42} & 36.4 & {\scriptsize 1.55} & 37.5$^{\dagger}$ & {\scriptsize 1.10} \\ \midrule

~~~Retrieved (closest) using \\

~~~~~~\sbert{} representations
& 1 & 4 & 24.5 & {\scriptsize 1.47} & 15.4 & {\scriptsize 1.67} & 18.9$^{\dagger}$ & {\scriptsize 1.56} & 31.1 & {\scriptsize 1.14} & 36.0 & {\scriptsize 1.41} & 33.4$^{\dagger}$ & {\scriptsize 1.05} \\
& 1 & 9 & 24.9 & {\scriptsize 1.10} & 15.9 & {\scriptsize 0.88} & 19.4$^{\dagger}$ & {\scriptsize 0.83} & 30.4 & {\scriptsize 1.09} & 31.1 & {\scriptsize 0.87} & 30.7$^{\dagger}$ & {\scriptsize 0.76} \\

~~~~~~~~~+ hybrid
& 1 & 4 & 25.9 & {\scriptsize 0.58} & 23.4 & {\scriptsize 1.31} & 24.6$^{\dagger}$ & {\scriptsize 0.81} & 39.3 & {\scriptsize 0.70} & 39.4 & {\scriptsize 1.48} & \textbf{39.3} & {\scriptsize 0.70} \\
& 1 & 9 & 26.6 & {\scriptsize 0.91} & 25.4 & {\scriptsize 0.79} & \textbf{26.0}$^{\dagger}$ & {\scriptsize 0.57} & 39.3 & {\scriptsize 0.80} & 35.7 & {\scriptsize 1.40} & {37.4}$^{\dagger}$ & {\scriptsize 0.86} \\ \addlinespace

~~~~~~Semantic rule representations
& 1 & 4 & 23.9 & {\scriptsize 0.48} & 23.3 & {\scriptsize 0.56} & 23.6$^{\dagger}$ & {\scriptsize 0.43} & 33.3 & {\scriptsize 1.15} & 38.6 & {\scriptsize 1.49} & 35.8$^{\dagger}$ & {\scriptsize 1.21} \\
& 1 & 9 & 22.9 & {\scriptsize 0.56} & 25.1 & {\scriptsize 0.68} & 23.9$^{\dagger}$ & {\scriptsize 0.50} & 33.0 & {\scriptsize 1.66} & 33.9 & {\scriptsize 2.08} & 33.4$^{\dagger}$ & {\scriptsize 1.76} \\

~~~~~~~~~+ hybrid
& 1 & 4 & 26.6 & {\scriptsize 0.87} & 26.7 & {\scriptsize 0.61} & 26.6 & {\scriptsize 0.54} & 39.4 & {\scriptsize 1.04} & 39.4 & {\scriptsize 1.27} & \textbf{39.4} & {\scriptsize {0.89}} \\
& 1 & 9 & 26.2 & {\scriptsize 0.60} & 28.9 & {\scriptsize 0.91} & \textbf{27.5} & {\scriptsize 0.48} & 40.0 & {\scriptsize 1.11} & 36.2 & {\scriptsize 1.12} & 38.0$^{\dagger}$ & {\scriptsize 0.82} \\

~~~~~~~~~and $k$-means clustering \\
~~~~~~~~~~~~w/ random cluster selection
& 1 & 4 & 22.8 & {\scriptsize 0.89} & 28.5 & {\scriptsize 1.21} & 25.3$^{\dagger}$ & {\scriptsize 0.96} & 29.5 & {\scriptsize 0.81} & 46.3 & {\scriptsize 1.18} & 36.0$^{\dagger}$ & {\scriptsize 0.83} \\
& 1 & 9 & 22.1 & {\scriptsize 0.75} & 30.6 & {\scriptsize 0.98} & 25.7$^{\dagger}$ & {\scriptsize 0.70} & 29.7 & {\scriptsize 0.89} & 45.1 & {\scriptsize 1.59} & 35.8$^{\dagger}$ & {\scriptsize 0.96} \\

~~~~~~~~~~~~~~~+ hybrid
& 1 & 4 & 25.7 & {\scriptsize 1.05} & 26.6 & {\scriptsize 0.68} & 26.1$^{\dagger}$ & {\scriptsize 0.70} & 37.9 & {\scriptsize 0.94} & 40.5 & {\scriptsize 1.64} & 39.1 & {\scriptsize 1.15} \\
& 1 & 9 & 26.4 & {\scriptsize 0.69} & 28.9 & {\scriptsize 1.88} & \textbf{27.6} & {\scriptsize {1.20}} & 38.7 & {\scriptsize 1.19} & 37.0 & {\scriptsize 1.36} & 37.8$^{\dagger}$ & {\scriptsize 0.84} \\

~~~~~~~~~~~~w/ closest cluster selection
& 1 & 4 & 22.1 & {\scriptsize 0.66} & 27.7 & {\scriptsize 1.32} & 24.6$^{\dagger}$ & {\scriptsize 0.90} & 30.1 & {\scriptsize 0.79} & 43.2 & {\scriptsize 1.82} & 35.5$^{\dagger}$ & {\scriptsize 1.05} \\
& 1 & 9 & 21.2 & {\scriptsize 0.39} & 28.9 & {\scriptsize 1.37} & 24.5$^{\dagger}$ & {\scriptsize 0.69} & 31.0 & {\scriptsize 0.96} & 41.5 & {\scriptsize 2.06} & 35.5$^{\dagger}$ & {\scriptsize 1.04} \\

~~~~~~~~~~~~~~~+ hybrid
& 1 & 4 & 25.3 & {\scriptsize 0.59} & 26.2 & {\scriptsize 0.82} & 25.8$^{\dagger}$ & {\scriptsize 0.49} & 38.2 & {\scriptsize 0.81} & 40.5 & {\scriptsize 1.44} & \textbf{39.3} & {\scriptsize 0.75} \\
& 1 & 9 & 25.5 & {\scriptsize 0.62} & 27.8 & {\scriptsize 1.30} & 26.6$^{*}$ & {\scriptsize 0.83} & 39.7 & {\scriptsize 1.22} & 37.5 & {\scriptsize 1.49} & 38.6$^{*}$ & {\scriptsize 0.97} \\

~~~~~~~~~and $k$-means++ clustering \\
~~~~~~~~~~~~w/ furthest cluster selection
& 1 & 4 & 22.7 & {\scriptsize 0.93} & 28.1 & {\scriptsize 1.72} & 25.1$^{\dagger}$ & {\scriptsize 1.24} & 28.4 & {\scriptsize 0.80} & {46.4} & {\scriptsize {1.58}} & 35.2$^{\dagger}$ & {\scriptsize 0.98} \\
& 1 & 9 & 21.6 & {\scriptsize 0.54} & 31.6 & {\scriptsize 1.11} & 25.7$^{\dagger}$ & {\scriptsize 0.73} & 29.1 & {\scriptsize 0.92} & 46.0 & {\scriptsize 1.78} & 35.7$^{\dagger}$ & {\scriptsize 1.09} \\

~~~~~~~~~~~~~~~+ hybrid
& 1 & 4 & 25.0 & {\scriptsize 0.65} & 26.1 & {\scriptsize 1.55} & 25.5$^{\dagger}$ & {\scriptsize 0.97} & 37.6 & {\scriptsize 0.84} & 40.2 & {\scriptsize 1.57} & 38.8 & {\scriptsize 0.89} \\
& 1 & 9 & 25.8 & {\scriptsize 0.38} & 28.9 & {\scriptsize 1.39} & \textbf{27.3} & {\scriptsize 0.79} & 39.1 & {\scriptsize 1.31} & 37.7 & {\scriptsize 1.67} & 38.4$^{\dagger}$ & {\scriptsize 1.19} \\

~~~~~~~~~~~~w/ closest cluster selection 
& 1 & 4 & 22.3 & {\scriptsize 0.50} & 26.3 & {\scriptsize 1.05} & 24.1$^{\dagger}$ & {\scriptsize 0.70} & 31.3 & {\scriptsize 1.12} & 42.7 & {\scriptsize 1.75} & 36.1$^{\dagger}$ & {\scriptsize 1.11} \\
& 1 & 9 & 21.7 & {\scriptsize 0.52} & 29.6 & {\scriptsize 0.72} & 25.0$^{\dagger}$ & {\scriptsize 0.36} & 31.4 & {\scriptsize 1.20} & 41.8 & {\scriptsize 2.37} & 35.8$^{\dagger}$ & {\scriptsize 1.46} \\

~~~~~~~~~~~~~~~+ hybrid
& 1 & 4 & 25.4 & {\scriptsize 0.82} & 26.3 & {\scriptsize 1.42} & 25.8$^{\dagger}$ & {\scriptsize 1.00} & 38.2 & {\scriptsize 1.17} & 40.2 & {\scriptsize 1.45} & \textbf{39.2} & {\scriptsize 1.04} \\
& 1 & 9 & 26.1 & {\scriptsize 0.46} & 27.9 & {\scriptsize 1.54} & 27.0 & {\scriptsize 0.95} & 39.8 & {\scriptsize 1.39} & 37.2 & {\scriptsize 1.71} & 38.5$^{*}$ & {\scriptsize 1.20} \\ \midrule

~~~Retrieved (closest) using \\
~~~~~~Semantic rule representations \\
~~~~~~~~~and $k$-means, \emph{summarization} \\
~~~~~~~~~~~~w/ random cluster selection
& 1 & 4 & 21.1 & {\scriptsize 0.77} & 33.9 & {\scriptsize 1.00} & 26.0$^{\dagger}$ & {\scriptsize 0.84} & 27.8 & {\scriptsize 0.60} & 44.3 & {\scriptsize 1.63} & 34.1$^{\dagger}$ & {\scriptsize 0.90} \\
& 1 & 9 & 20.8 & {\scriptsize 0.84} & 41.1 & {\scriptsize 1.25} & \textbf{27.6} & {\scriptsize {1.01}} & 28.4 & {\scriptsize 1.13} & 41.0 & {\scriptsize 1.96} & 33.6$^{\dagger}$ & {\scriptsize 1.36} \\

~~~~~~~~~~~~~~~+ hybrid
& 1 & 4 & 22.8 & {\scriptsize 0.29} & 29.1 & {\scriptsize 0.91} & 25.5$^{\dagger}$ & {\scriptsize 0.38} & 32.8 & {\scriptsize 0.96} & 41.3 & {\scriptsize 1.46} & \textbf{36.6}$^{\dagger}$ & {\scriptsize 1.08} \\
& 1 & 9 & 23.6 & {\scriptsize 0.69} & 33.2 & {\scriptsize 2.10} & \textbf{27.6} & {\scriptsize {1.18}} & 35.1 & {\scriptsize 1.45} & 35.6 & {\scriptsize 1.54} & 35.4$^{\dagger}$ & {\scriptsize 1.32} \\

~~~~~~~~~~~~w/ closest cluster selection
& 1 & 4 & 21.3 & {\scriptsize 0.74} & 32.6 & {\scriptsize 1.58} & 25.8$^{\dagger}$ & {\scriptsize 0.96} & 28.6 & {\scriptsize 0.86} & 40.7 & {\scriptsize 1.53} & 33.6$^{\dagger}$ & {\scriptsize 0.93} \\
& 1 & 9 & 20.2 & {\scriptsize 0.71} & 41.6 & {\scriptsize 1.61} & 27.2 & {\scriptsize 0.91} & 29.8 & {\scriptsize 1.02} & 39.0 & {\scriptsize 1.36} & 33.8$^{\dagger}$ & {\scriptsize 1.00} \\

~~~~~~~~~~~~~~~+ hybrid
& 1 & 4 & 23.5 & {\scriptsize 0.56} & 29.3 & {\scriptsize 1.45} & 26.1$^{\dagger}$ & {\scriptsize 0.85} & 32.7 & {\scriptsize 1.70} & 40.7 & {\scriptsize 1.33} & 36.2$^{\dagger}$ & {\scriptsize 1.47} \\
& 1 & 9 & 22.7 & {\scriptsize 0.40} & 33.1 & {\scriptsize 1.69} & 27.0 & {\scriptsize 0.77} & 34.8 & {\scriptsize 1.22} & 35.6 & {\scriptsize 1.55} & 35.2$^{\dagger}$ & {\scriptsize 1.06} \\

~~~~~~~~~and $k$-means++, \emph{summarization} \\
~~~~~~~~~~~~w/ furthest cluster selection
& 1 & 4 & 22.1 & {\scriptsize 0.57} & 35.2 & {\scriptsize 1.25} & 27.1 & {\scriptsize 0.74} & 28.0 & {\scriptsize 0.90} & 44.4 & {\scriptsize 2.00} & 34.4$^{\dagger}$ & {\scriptsize 1.20} \\
& 1 & 9 & 20.0 & {\scriptsize 0.68} & {42.4} & {\scriptsize {1.38}} & \textbf{27.2} & {\scriptsize 0.88} & 29.4 & {\scriptsize 1.46} & 42.0 & {\scriptsize 1.85} & 34.6$^{\dagger}$ & {\scriptsize 1.51} \\

~~~~~~~~~~~~~~~+ hybrid
& 1 & 4 & 23.2 & {\scriptsize 0.84} & 29.2 & {\scriptsize 1.31} & 25.9$^{\dagger}$ & {\scriptsize 0.90} & 33.0 & {\scriptsize 1.39} & 41.7 & {\scriptsize 1.27} & \textbf{36.8}$^{\dagger}$ & {\scriptsize 1.16} \\
& 1 & 9 & 23.2 & {\scriptsize 0.38} & 32.9 & {\scriptsize 1.95} & \textbf{27.2} & {\scriptsize 0.89} & 35.4 & {\scriptsize 1.40} & 36.2 & {\scriptsize 1.23} & 35.8$^{\dagger}$ & {\scriptsize 1.01} \\

~~~~~~~~~~~~w/ closest cluster selection
& 1 & 4 & 21.0 & {\scriptsize 1.27} & 32.1 & {\scriptsize 1.95} & 25.4$^{\dagger}$ & {\scriptsize 1.51} & 29.5 & {\scriptsize 1.01} & 41.1 & {\scriptsize 1.58} & 34.3$^{\dagger}$ & {\scriptsize 0.99} \\
& 1 & 9 & 19.4 & {\scriptsize 0.68} & 40.3 & {\scriptsize 1.53} & 26.2$^{\dagger}$ & {\scriptsize 0.91} & 29.3 & {\scriptsize 0.76} & 37.8 & {\scriptsize 1.48} & 33.0$^{\dagger}$ & {\scriptsize 0.80} \\

~~~~~~~~~~~~~~~+ hybrid
& 1 & 4 & 22.4 & {\scriptsize 0.69} & 28.9 & {\scriptsize 1.96} & 25.2$^{\dagger}$ & {\scriptsize 1.12} & 32.5 & {\scriptsize 0.98} & 40.9 & {\scriptsize 1.60} & 36.2$^{\dagger}$ & {\scriptsize 1.05} \\
& 1 & 9 & 22.5 & {\scriptsize 0.47} & 32.2 & {\scriptsize 1.78} & 26.5$^{*}$ & {\scriptsize 0.81} & 35.6 & {\scriptsize 1.13} & 35.4 & {\scriptsize 1.41} & 35.5$^{\dagger}$ & {\scriptsize 1.07} \\ \bottomrule
                                \end{tabular}

\caption{
Full results obtained with \tacred{} using \qwensmall{} and \qwenlarge{}.
This table complements Table \ref{t:results_fs-tacred_main}
by providing results with alternative cluster selection strategies and summarizing examples.
We observe the same trends as in Table \ref{t:results_fs-tacred_main}: our hybrid approach
combining additional examples generated with an LLM and retrieved with semantic rule representations outperforms the alternatives.
$^{*}$ and $^{\dagger}$ indicate the best method in the whole column is statistically significantly better at $p < 0.05$ and $p < 0.01$, respectively, 
  based on one-sided paired bootstrap test with replacement on F1 scores.
}
  \label{t:results_fs-tacred_full_qwen}
\end{table*}

\begin{table*}
  \centering
  \small
  \setlength{\tabcolsep}{.061in}
  \setlength{\tabcolsep}{0.055in}
\begin{tabular}{l r@{+}l
                r@{ \scriptsize$\pm$ }l r@{ \scriptsize$\pm$ }l l@{ \scriptsize$\pm$ }l
                r@{ \scriptsize$\pm$ }l r@{ \scriptsize$\pm$ }l l@{ \scriptsize$\pm$ }l}
\toprule
  &
  \multicolumn{2}{c}{\#ex.} &
  \multicolumn{6}{c}{\qwensmall} &
  \multicolumn{6}{c}{\qwenlarge} 
  \\ \cmidrule(lr){4-9} \cmidrule(lr){10-15}
& G & A &
  \multicolumn{2}{c}{P} & \multicolumn{2}{c}{R} & \multicolumn{2}{c}{F1} &
  \multicolumn{2}{c}{P} & \multicolumn{2}{c}{R} & \multicolumn{2}{c}{F1} \\ \midrule

Few-shot baseline & 1 & 0 & 41.7 & {\scriptsize 1.37} & 26.4 & {\scriptsize 1.17} & 32.4$^{\dagger}$ & {\scriptsize 1.14} & 43.5 & {\scriptsize 0.81} & 42.3 & {\scriptsize 0.73} & 42.9$^{\dagger}$ & {\scriptsize 0.52} \\ \midrule
Including \emph{A}dditional examples \\

~~~Paraphrased with LLM  
& 1 & 4 & {44.3} & {\scriptsize {1.91}} & 13.4 & {\scriptsize 0.78} & 20.5$^{\dagger}$ & {\scriptsize 1.05} & 42.6 & {\scriptsize 1.17} & {48.8} & {\scriptsize {0.75}} & \textbf{45.5} & {\scriptsize {0.75}} \\
& 1 & 9 & 40.2 & {\scriptsize 2.37} & 13.3 & {\scriptsize 0.97} & 20.0$^{\dagger}$ & {\scriptsize 1.36} & 45.6 & {\scriptsize 0.72} & 43.9 & {\scriptsize 0.51} & 44.7$^{\dagger}$ & {\scriptsize 0.09} \\ 

~~~Generated with LLM
& 1 & 4 & 43.8 & {\scriptsize 0.95} & 29.4 & {\scriptsize 0.75} & \textbf{35.2}$^{\dagger}$ & {\scriptsize 0.75} & 42.7 & {\scriptsize 1.20} & {48.8} & {\scriptsize {0.76}} & \textbf{45.5} & {\scriptsize {0.78}} \\
& 1 & 9 & 42.0 & {\scriptsize 0.96} & 28.4 & {\scriptsize 0.68} & 33.9$^{\dagger}$ & {\scriptsize 0.79} & {45.8} & {\scriptsize {0.92}} & 44.2 & {\scriptsize 0.37} & 45.0 & {\scriptsize 0.31} \\ \midrule

~~~Retrieved (closest) using \\

~~~~~~\sbert{} representations
& 1 & 4 & 43.3 & {\scriptsize 0.94} & 22.4 & {\scriptsize 0.48} & 29.5$^{\dagger}$ & {\scriptsize 0.57} & 43.7 & {\scriptsize 0.95} & 43.0 & {\scriptsize 1.86} & 43.4$^{\dagger}$ & {\scriptsize 1.13} \\
& 1 & 9 & 42.4 & {\scriptsize 1.10} & 21.2 & {\scriptsize 0.45} & 28.2$^{\dagger}$ & {\scriptsize 0.47} & 44.1 & {\scriptsize 1.04} & 39.8 & {\scriptsize 1.52} & 41.8$^{\dagger}$ & {\scriptsize 1.08} \\

~~~~~~~~~+ hybrid
& 1 & 4 & 44.1 & {\scriptsize 1.19} & 29.4 & {\scriptsize 0.68} & \textbf{35.3}$^{\dagger}$ & {\scriptsize 0.86} & 44.0 & {\scriptsize 0.93} & 43.1 & {\scriptsize 1.77} & \textbf{43.5}$^{\dagger}$ & {\scriptsize 1.13} \\
& 1 & 9 & 43.0 & {\scriptsize 0.73} & 29.2 & {\scriptsize 0.51} & 34.7$^{\dagger}$ & {\scriptsize 0.50} & 44.7 & {\scriptsize 1.18} & 39.9 & {\scriptsize 1.53} & 42.1$^{\dagger}$ & {\scriptsize 1.00} \\ \addlinespace

~~~~~~Semantic rule representations
& 1 & 4 & 40.6 & {\scriptsize 1.44} & 33.0 & {\scriptsize 1.30} & 36.4$^{\dagger}$ & {\scriptsize 1.37} & 43.3 & {\scriptsize 0.66} & 45.6 & {\scriptsize 1.63} & \textbf{44.4}$^{\dagger}$ & {\scriptsize 0.98} \\
& 1 & 9 & 38.1 & {\scriptsize 1.12} & 33.0 & {\scriptsize 0.57} & 35.3$^{\dagger}$ & {\scriptsize 0.78} & 42.7 & {\scriptsize 0.66} & 42.5 & {\scriptsize 1.63} & 42.6$^{\dagger}$ & {\scriptsize 0.98} \\

~~~~~~~~~+ hybrid
& 1 & 4 & 44.1 & {\scriptsize 1.50} & 31.6 & {\scriptsize 1.20} & \textbf{36.8}$^{\dagger}$ & {\scriptsize 1.25} & 43.2 & {\scriptsize 0.61} & 45.4 & {\scriptsize 1.49} & 44.2$^{\dagger}$ & {\scriptsize 0.86} \\
& 1 & 9 & 41.9 & {\scriptsize 1.36} & 31.6 & {\scriptsize 0.90} & 36.0$^{\dagger}$ & {\scriptsize 1.06} & 42.8 & {\scriptsize 1.22} & 43.0 & {\scriptsize 1.72} & 42.9$^{\dagger}$ & {\scriptsize 1.19} \\

~~~~~~~~~and $k$-means clustering \\
~~~~~~~~~~~~w/ random cluster selection
& 1 & 4 & 39.1 & {\scriptsize 1.66} & 33.0 & {\scriptsize 0.98} & 35.8$^{\dagger}$ & {\scriptsize 1.22} & 40.3 & {\scriptsize 0.97} & 47.1 & {\scriptsize 1.71} & 43.4$^{\dagger}$ & {\scriptsize 1.03} \\
& 1 & 9 & 35.6 & {\scriptsize 1.23} & 32.4 & {\scriptsize 0.58} & 33.9$^{\dagger}$ & {\scriptsize 0.85} & 39.9 & {\scriptsize 0.99} & 46.3 & {\scriptsize 1.27} & 42.9$^{\dagger}$ & {\scriptsize 0.88} \\

~~~~~~~~~~~~~~~+ hybrid
& 1 & 4 & 43.9 & {\scriptsize 1.49} & 31.6 & {\scriptsize 0.99} & 36.7$^{\dagger}$ & {\scriptsize 1.09} & 40.3 & {\scriptsize 1.14} & 46.7 & {\scriptsize 1.86} & 43.2$^{\dagger}$ & {\scriptsize 1.21} \\
& 1 & 9 & 43.0 & {\scriptsize 1.46} & 32.2 & {\scriptsize 1.29} & \textbf{36.8}$^{\dagger}$ & {\scriptsize 1.35} & 39.9 & {\scriptsize 0.82} & 45.9 & {\scriptsize 0.81} & 42.7$^{\dagger}$ & {\scriptsize 0.64} \\

~~~~~~~~~~~~w/ closest cluster selection
& 1 & 4 & 38.1 & {\scriptsize 1.09} & 34.1 & {\scriptsize 0.61} & 36.0$^{\dagger}$ & {\scriptsize 0.81} & 41.1 & {\scriptsize 0.94} & 46.5 & {\scriptsize 1.71} & 43.6$^{\dagger}$ & {\scriptsize 1.12} \\
& 1 & 9 & 36.1 & {\scriptsize 1.18} & 33.7 & {\scriptsize 0.71} & 34.9$^{\dagger}$ & {\scriptsize 0.91} & 40.4 & {\scriptsize 0.89} & 45.4 & {\scriptsize 1.68} & 42.7$^{\dagger}$ & {\scriptsize 1.15} \\

~~~~~~~~~~~~~~~+ hybrid
& 1 & 4 & 43.7 & {\scriptsize 1.28} & 31.8 & {\scriptsize 0.72} & \textbf{36.8}$^{\dagger}$ & {\scriptsize 0.84} & 41.3 & {\scriptsize 1.05} & 46.4 & {\scriptsize 1.59} & \textbf{43.7}$^{\dagger}$ & {\scriptsize 1.13} \\
& 1 & 9 & 42.8 & {\scriptsize 1.06} & 32.1 & {\scriptsize 0.81} & 36.6$^{\dagger}$ & {\scriptsize 0.81} & 40.1 & {\scriptsize 1.29} & 44.8 & {\scriptsize 1.62} & 42.3$^{\dagger}$ & {\scriptsize 1.29} \\

~~~~~~~~~and $k$-means++ clustering \\
~~~~~~~~~~~~w/ furthest cluster selection
& 1 & 4 & 40.0 & {\scriptsize 1.20} & 33.7 & {\scriptsize 1.32} & 36.6$^{\dagger}$ & {\scriptsize 1.16} & 39.6 & {\scriptsize 0.85} & 47.9 & {\scriptsize 1.72} & 43.3$^{\dagger}$ & {\scriptsize 0.99} \\
& 1 & 9 & 37.2 & {\scriptsize 0.98} & 33.5 & {\scriptsize 0.64} & 35.3$^{\dagger}$ & {\scriptsize 0.74} & 39.6 & {\scriptsize 0.68} & 46.1 & {\scriptsize 1.20} & 42.6$^{\dagger}$ & {\scriptsize 0.63} \\

~~~~~~~~~~~~~~~+ hybrid
& 1 & 4 & 43.4 & {\scriptsize 1.49} & 31.1 & {\scriptsize 0.68} & 36.2$^{\dagger}$ & {\scriptsize 0.91} & 39.8 & {\scriptsize 0.49} & 47.7 & {\scriptsize 1.48} & 43.4$^{\dagger}$ & {\scriptsize 0.68} \\
& 1 & 9 & 42.5 & {\scriptsize 1.35} & 31.3 & {\scriptsize 0.94} & 36.1$^{\dagger}$ & {\scriptsize 1.00} & 39.2 & {\scriptsize 0.86} & 45.9 & {\scriptsize 1.04} & 42.3$^{\dagger}$ & {\scriptsize 0.80} \\

~~~~~~~~~~~~w/ closest cluster selection
& 1 & 4 & 39.1 & {\scriptsize 1.16} & 34.0 & {\scriptsize 0.89} & 36.4$^{\dagger}$ & {\scriptsize 0.96} & 40.9 & {\scriptsize 1.15} & 46.3 & {\scriptsize 1.36} & 43.5$^{\dagger}$ & {\scriptsize 1.06} \\
& 1 & 9 & 36.2 & {\scriptsize 1.20} & 33.1 & {\scriptsize 0.77} & 34.6$^{\dagger}$ & {\scriptsize 0.94} & 40.1 & {\scriptsize 1.15} & 44.6 & {\scriptsize 1.33} & 42.2$^{\dagger}$ & {\scriptsize 1.06} \\

~~~~~~~~~~~~~~~+ hybrid
& 1 & 4 & 43.5 & {\scriptsize 1.16} & 31.9 & {\scriptsize 0.65} & \textbf{36.8}$^{\dagger}$ & {\scriptsize 0.68} & 41.1 & {\scriptsize 1.12} & 46.6 & {\scriptsize 1.27} & \textbf{43.7}$^{\dagger}$ & {\scriptsize 0.97} \\
& 1 & 9 & 42.7 & {\scriptsize 0.88} & 31.7 & {\scriptsize 0.57} & 36.3$^{\dagger}$ & {\scriptsize 0.54} & 40.2 & {\scriptsize 1.15} & 44.7 & {\scriptsize 1.04} & 42.4$^{\dagger}$ & {\scriptsize 0.95} \\ \midrule

~~~Retrieved (closest) using \\
~~~~~~Semantic rule representations \\
~~~~~~~~~and $k$-means, \emph{summarization} \\
~~~~~~~~~~~~w/ random cluster selection
& 1 & 4 & 36.5 & {\scriptsize 0.97} & 36.0 & {\scriptsize 0.78} & 36.2$^{\dagger}$ & {\scriptsize 0.82} & 39.5 & {\scriptsize 0.89} & 44.7 & {\scriptsize 1.24} & 42.0$^{\dagger}$ & {\scriptsize 0.85} \\
& 1 & 9 & 33.1 & {\scriptsize 0.56} & 39.5 & {\scriptsize 0.58} & 36.0$^{\dagger}$ & {\scriptsize 0.46} & 38.4 & {\scriptsize 1.34} & 42.5 & {\scriptsize 0.85} & 40.3$^{\dagger}$ & {\scriptsize 1.05} \\

~~~~~~~~~~~~~~~+ hybrid 
& 1 & 4 & 41.2 & {\scriptsize 0.72} & 35.8 & {\scriptsize 0.68} & 38.3 & {\scriptsize 0.64} & 39.9 & {\scriptsize 0.76} & 45.7 & {\scriptsize 1.23} & \textbf{42.6}$^{\dagger}$ & {\scriptsize 0.66} \\
& 1 & 9 & 38.8 & {\scriptsize 0.94} & 36.8 & {\scriptsize 0.77} & 37.7$^{*}$ & {\scriptsize 0.78} & 39.0 & {\scriptsize 1.36} & 43.5 & {\scriptsize 0.91} & 41.1$^{\dagger}$ & {\scriptsize 1.07} \\

~~~~~~~~~~~~w/ closest cluster selection
& 1 & 4 & 36.1 & {\scriptsize 0.84} & 36.3 & {\scriptsize 0.90} & 36.2$^{\dagger}$ & {\scriptsize 0.75} & 39.7 & {\scriptsize 0.90} & 44.4 & {\scriptsize 1.07} & 41.9$^{\dagger}$ & {\scriptsize 0.80} \\
& 1 & 9 & 33.1 & {\scriptsize 1.24} & {39.9} & {\scriptsize {1.06}} & 36.2$^{\dagger}$ & {\scriptsize 1.12} & 39.4 & {\scriptsize 0.60} & 42.4 & {\scriptsize 0.74} & 40.8$^{\dagger}$ & {\scriptsize 0.42} \\

~~~~~~~~~~~~~~~+ hybrid
& 1 & 4 & 41.2 & {\scriptsize 0.77} & 36.0 & {\scriptsize 0.74} & \textbf{38.4} & {\scriptsize {0.66}} & 40.1 & {\scriptsize 0.83} & 44.6 & {\scriptsize 1.33} & 42.2$^{\dagger}$ & {\scriptsize 0.83} \\
& 1 & 9 & 39.2 & {\scriptsize 1.05} & 37.3 & {\scriptsize 1.17} & 38.2 & {\scriptsize 1.03} & 39.6 & {\scriptsize 0.95} & 42.8 & {\scriptsize 1.66} & 41.1$^{\dagger}$ & {\scriptsize 1.14} \\

~~~~~~~~~and $k$-means++, \emph{summarization} \\
~~~~~~~~~~~~w/ furthest cluster selection
& 1 & 4 & 37.8 & {\scriptsize 1.19} & 38.0 & {\scriptsize 1.23} & 37.9 & {\scriptsize 1.09} & 39.3 & {\scriptsize 0.74} & 45.8 & {\scriptsize 1.03} & 42.3$^{\dagger}$ & {\scriptsize 0.59} \\
& 1 & 9 & 33.2 & {\scriptsize 1.31} & 38.4 & {\scriptsize 1.20} & 35.6$^{\dagger}$ & {\scriptsize 1.18} & 38.5 & {\scriptsize 1.12} & 43.2 & {\scriptsize 0.79} & 40.7$^{\dagger}$ & {\scriptsize 0.91} \\

~~~~~~~~~~~~~~~+ hybrid
& 1 & 4 & 41.1 & {\scriptsize 1.40} & 35.8 & {\scriptsize 1.76} & \textbf{38.3} & {\scriptsize 1.50} & 39.0 & {\scriptsize 1.05} & 45.4 & {\scriptsize 0.39} & 41.9$^{\dagger}$ & {\scriptsize 0.69} \\
& 1 & 9 & 39.3 & {\scriptsize 1.10} & 36.7 & {\scriptsize 0.95} & 37.9 & {\scriptsize 0.91} & 38.5 & {\scriptsize 1.06} & 43.3 & {\scriptsize 1.08} & 40.7$^{\dagger}$ & {\scriptsize 0.96} \\

~~~~~~~~~~~~w/ closest cluster selection
& 1 & 4 & 37.3 & {\scriptsize 1.16} & 37.9 & {\scriptsize 1.19} & 37.6$^{\dagger}$ & {\scriptsize 1.03} & 40.6 & {\scriptsize 1.29} & 44.5 & {\scriptsize 1.11} & \textbf{42.4}$^{\dagger}$ & {\scriptsize 1.05} \\
& 1 & 9 & 33.0 & {\scriptsize 1.02} & 39.8 & {\scriptsize 0.93} & 36.1$^{\dagger}$ & {\scriptsize 0.86} & 39.6 & {\scriptsize 1.14} & 42.6 & {\scriptsize 0.82} & 41.0$^{\dagger}$ & {\scriptsize 0.89} \\

~~~~~~~~~~~~~~~+ hybrid
& 1 & 4 & 40.9 & {\scriptsize 0.92} & 35.9 & {\scriptsize 0.77} & 38.2 & {\scriptsize 0.72} & 40.3 & {\scriptsize 1.18} & 44.6 & {\scriptsize 1.28} & 42.3$^{\dagger}$ & {\scriptsize 1.09} \\
& 1 & 9 & 39.1 & {\scriptsize 1.11} & 37.1 & {\scriptsize 0.83} & 38.1 & {\scriptsize 0.83} & 39.3 & {\scriptsize 1.09} & 42.5 & {\scriptsize 0.65} & 40.8$^{\dagger}$ & {\scriptsize 0.78} \\ \bottomrule
\end{tabular}
\caption{
Full results obtained with \fewrel{} using \qwensmall{} and \qwenlarge{}.
This table complements Table \ref{t:results_fs-fewrel_main}
by providing results with alternative cluster selection strategies and summarizing examples.
We observe the same trends as in Table \ref{t:results_fs-fewrel_main}: our hybrid approach
combining additional examples generated with an LLM and retrieved with semantic rule representations outperforms the alternatives with \qwensmall.
$^{*}$ and $^{\dagger}$ indicate the best method in the whole column is statistically significantly better at $p < 0.05$ and $p < 0.01$, respectively, 
  based on one-sided paired bootstrap test with replacement on F1 scores.
}
  \label{t:results_fs-fewrel_full_qwen}
\end{table*}

\begin{table*}
  \centering
  \small
  \setlength{\tabcolsep}{.061in}
\setlength{\tabcolsep}{.061in}
\setlength{\tabcolsep}{0.055in}
\begin{tabular}{l r@{+}l
                r@{ \scriptsize$\pm$ }l r@{ \scriptsize$\pm$ }l l@{ \scriptsize$\pm$ }l
                r@{ \scriptsize$\pm$ }l r@{ \scriptsize$\pm$ }l l@{ \scriptsize$\pm$ }l}
\toprule
  &
  \multicolumn{2}{c}{\#ex.} &
  \multicolumn{6}{c}{Gemma3-4B} &
  \multicolumn{6}{c}{Gemma3-12B}
  \\ \cmidrule(lr){4-9} \cmidrule(lr){10-15}
& G & A &
  \multicolumn{2}{c}{P} & \multicolumn{2}{c}{R} & \multicolumn{2}{c}{F1} &
  \multicolumn{2}{c}{P} & \multicolumn{2}{c}{R} & \multicolumn{2}{c}{F1} \\ \midrule
Few-shot baseline & 1 & 0 & {11.9} & {\scriptsize {0.60}} & 18.5 & {\scriptsize 0.54} & 14.5 & {\scriptsize 0.58} & {25.3} & {\scriptsize {0.41}} & 51.3 & {\scriptsize 1.69} & {33.9} & {\scriptsize {0.62}} \\ \midrule
Including \emph{A}dditional examples \\

~~~Paraphrased with LLM  
& 1 & 4 & 08.9 & {\scriptsize 0.49} & 15.7 & {\scriptsize 1.02} & 11.3$^{\dagger}$ & {\scriptsize 0.64} & 28.6 & {\scriptsize 1.71} & 33.1 & {\scriptsize 1.81} & 30.7$^{\dagger}$ & {\scriptsize 1.68} \\ 
& 1 & 9 & 07.9 & {\scriptsize 0.59} & 14.0 & {\scriptsize 1.39} & 10.1$^{\dagger}$ & {\scriptsize 0.83} & 29.0 & {\scriptsize 1.58} & 30.0 & {\scriptsize 2.06} & 29.5$^{\dagger}$ & {\scriptsize 1.77} \\ 

~~~Generated with LLM
& 1 & 4 & 08.9 & {\scriptsize 0.30} & 38.1 & {\scriptsize 1.43} & \textbf{14.5}$^{\dagger}$ & {\scriptsize 0.48} & 21.4 & {\scriptsize 0.35} & 62.6 & {\scriptsize 2.13} & \textbf{31.9}$^{\dagger}$ & {\scriptsize 0.59} \\ 
& 1 & 9 & 09.4 & {\scriptsize 0.41} & 31.5 & {\scriptsize 1.86} & \textbf{14.5}$^{\dagger}$ & {\scriptsize 0.65} & 20.8 & {\scriptsize 0.22} & 64.1 & {\scriptsize 1.78} & 31.4$^{\dagger}$ & {\scriptsize 0.43} \\ \midrule

~~~Retrieved (closest) using \\

~~~~~~\sbert{} representations
& 1 & 4 & 10.0 & {\scriptsize 0.53} & 27.8 & {\scriptsize 1.28} & 14.7 & {\scriptsize 0.72} & 23.5 & {\scriptsize 0.69} & 49.1 & {\scriptsize 1.89} & \textbf{31.8}$^{\dagger}$ & {\scriptsize 0.96} \\ 
& 1 & 9 & 10.1 & {\scriptsize 0.49} & 24.3 & {\scriptsize 0.97} & 14.3$^{\dagger}$ & {\scriptsize 0.63} & 23.3 & {\scriptsize 0.69} & 43.1 & {\scriptsize 1.23} & 30.2$^{\dagger}$ & {\scriptsize 0.85} \\

~~~~~~~~~+ hybrid
& 1 & 4 & 09.4 & {\scriptsize 0.33} & 38.0 & {\scriptsize 1.47} & \textbf{15.1} & {\scriptsize {0.53}} & 21.3 & {\scriptsize 0.37} & 61.9 & {\scriptsize 1.70} & 31.7$^{\dagger}$ & {\scriptsize 0.48} \\
& 1 & 9 & 09.7 & {\scriptsize 0.34} & 32.1 & {\scriptsize 1.51} & 14.9 & {\scriptsize 0.55} & 21.2 & {\scriptsize 0.33} & 62.0 & {\scriptsize 1.78} & 31.6$^{\dagger}$ & {\scriptsize 0.49} \\ \addlinespace

~~~~~~Semantic rule representations
& 1 & 4 & 08.9 & {\scriptsize 0.20} & 39.0 & {\scriptsize 1.15} & 14.5$^{*}$ & {\scriptsize 0.31} & 20.2 & {\scriptsize 0.44} & 60.7 & {\scriptsize 2.42} & 30.3$^{\dagger}$ & {\scriptsize 0.79} \\ 
& 1 & 9 & 08.6 & {\scriptsize 0.21} & 37.6 & {\scriptsize 1.53} & 14.0$^{\dagger}$ & {\scriptsize 0.38} & 19.6 & {\scriptsize 0.27} & 57.7 & {\scriptsize 1.59} & 29.3$^{\dagger}$ & {\scriptsize 0.48} \\

~~~~~~~~~+ hybrid
& 1 & 4 & 08.9 & {\scriptsize 0.39} & 40.6 & {\scriptsize 1.45} & 14.5$^{*}$ & {\scriptsize 0.60} & 21.2 & {\scriptsize 0.31} & 63.4 & {\scriptsize 2.26} & \textbf{31.8}$^{\dagger}$ & {\scriptsize 0.58} \\
& 1 & 9 & 09.3 & {\scriptsize 0.15} & 36.8 & {\scriptsize 1.49} & \textbf{14.8} & {\scriptsize 0.30} & 21.0 & {\scriptsize 0.49} & 64.4 & {\scriptsize 2.61} & 31.6$^{\dagger}$ & {\scriptsize 0.81} \\

~~~~~~~~~and $k$-means clustering \\
~~~~~~~~~~~~w/ random cluster selection
& 1 & 4 & 08.8 & {\scriptsize 0.35} & {42.8} & {\scriptsize {1.50}} & 14.5$^{*}$ & {\scriptsize 0.56} & 19.1 & {\scriptsize 0.30} & 62.2 & {\scriptsize 2.01} & 29.2$^{\dagger}$ & {\scriptsize 0.55} \\ 
& 1 & 9 & 08.8 & {\scriptsize 0.24} & 41.2 & {\scriptsize 1.42} & 14.5$^{*}$ & {\scriptsize 0.40} & 18.8 & {\scriptsize 0.37} & 61.4 & {\scriptsize 1.47} & 28.8$^{\dagger}$ & {\scriptsize 0.57} \\

~~~~~~~~~~~~~~~+ hybrid
& 1 & 4 & 08.8 & {\scriptsize 0.43} & 41.9 & {\scriptsize 1.90} & 14.6$^{*}$ & {\scriptsize 0.69} & 20.7 & {\scriptsize 0.30} & 62.7 & {\scriptsize 1.82} & 31.1$^{\dagger}$ & {\scriptsize 0.49} \\
& 1 & 9 & 09.1 & {\scriptsize 0.45} & 36.2 & {\scriptsize 1.69} & 14.6$^{*}$ & {\scriptsize 0.69} & 21.0 & {\scriptsize 0.27} & {65.2} & {\scriptsize {2.11}} & \textbf{31.7}$^{\dagger}$ & {\scriptsize 0.50} \\

~~~~~~~~~~~~w/ closest cluster selection
& 1 & 4 & 08.5 & {\scriptsize 0.34} & 42.5 & {\scriptsize 1.35} & 14.2$^{\dagger}$ & {\scriptsize 0.54} & 20.1 & {\scriptsize 0.34} & 61.5 & {\scriptsize 1.81} & 30.3$^{\dagger}$ & {\scriptsize 0.56} \\
& 1 & 9 & 08.8 & {\scriptsize 0.20} & 42.0 & {\scriptsize 1.23} & 14.6$^{*}$ & {\scriptsize 0.32} & 19.2 & {\scriptsize 0.26} & 60.0 & {\scriptsize 1.08} & 29.1$^{\dagger}$ & {\scriptsize 0.38} \\

~~~~~~~~~~~~~~~+ hybrid
& 1 & 4 & 08.8 & {\scriptsize 0.15} & 41.0 & {\scriptsize 0.78} & 14.4$^{*}$ & {\scriptsize 0.21} & 20.9 & {\scriptsize 0.30} & 63.6 & {\scriptsize 1.63} & 31.4$^{\dagger}$ & {\scriptsize 0.37} \\
& 1 & 9 & 09.5 & {\scriptsize 0.34} & 37.3 & {\scriptsize 1.39} &  \textbf{15.1} & {\scriptsize {0.53}} & 20.7 & {\scriptsize 0.29} & 64.6 & {\scriptsize 1.98} & 31.4$^{\dagger}$ & {\scriptsize 0.45} \\

~~~~~~~~~and $k$-means++ clustering \\
~~~~~~~~~~~~w/ furthest cluster selection
& 1 & 4 & 08.8 & {\scriptsize 0.40} & 42.7 & {\scriptsize 1.26} & 14.6$^{*}$ & {\scriptsize 0.61} & 19.7 & {\scriptsize 0.30} & 62.5 & {\scriptsize 1.94} & 30.0$^{\dagger}$ & {\scriptsize 0.55} \\ 
& 1 & 9 & 08.6 & {\scriptsize 0.14} & 39.7 & {\scriptsize 1.50} & 14.1$^{\dagger}$ & {\scriptsize 0.26} & 18.8 & {\scriptsize 0.34} & 61.2 & {\scriptsize 0.90} & 28.7$^{\dagger}$ & {\scriptsize 0.42} \\ 

~~~~~~~~~~~~~~~+ hybrid
& 1 & 4 & 08.9 & {\scriptsize 0.48} & 41.7 & {\scriptsize 1.72} & \textbf{14.7} & {\scriptsize 0.74} & 21.0 & {\scriptsize 0.23} & 63.0 & {\scriptsize 1.65} & 31.5$^{\dagger}$ & {\scriptsize 0.32} \\
& 1 & 9 & 09.1 & {\scriptsize 0.34} & 36.8 & {\scriptsize 1.92} & 14.6$^{*}$ & {\scriptsize 0.56} & 21.0 & {\scriptsize 0.38} & 64.6 & {\scriptsize 2.05} & \textbf{31.7}$^{\dagger}$ & {\scriptsize 0.53} \\

~~~~~~~~~~~~w/ closest cluster selection
& 1 & 4 & 08.7 & {\scriptsize 0.24} & 41.5 & {\scriptsize 1.63} & 14.4$^{\dagger}$ & {\scriptsize 0.41} & 19.9 & {\scriptsize 0.21} & 61.0 & {\scriptsize 1.79} & 30.0$^{\dagger}$ & {\scriptsize 0.41} \\
& 1 & 9 & 08.6 & {\scriptsize 0.35} & 40.0 & {\scriptsize 1.43} & 14.1$^{\dagger}$ & {\scriptsize 0.56} & 19.8 & {\scriptsize 0.23} & 60.5 & {\scriptsize 1.32} & 29.9$^{\dagger}$ & {\scriptsize 0.37} \\ 

~~~~~~~~~~~~~~~+ hybrid
& 1 & 4 & 08.9 & {\scriptsize 0.31} & 42.0 & {\scriptsize 1.68} & \textbf{14.7}$^{*}$ & {\scriptsize 0.50} & 20.9 & {\scriptsize 0.29} & 63.0 & {\scriptsize 2.06} & 31.3$^{\dagger}$ & {\scriptsize 0.48} \\
& 1 & 9 & 09.0 & {\scriptsize 0.27} & 37.1 & {\scriptsize 1.30} & 14.5$^{\dagger}$ & {\scriptsize 0.43} & 20.8 & {\scriptsize 0.26} & 64.4 & {\scriptsize 1.70} & 31.4$^{\dagger}$ & {\scriptsize 0.35} \\ \midrule

~~~Retrieved (closest) using \\
~~~~~~Semantic rule representations \\
~~~~~~~~~and $k$-means, \emph{summarization} \\
~~~~~~~~~~~~w/ random cluster selection
& 1 & 4 & 08.0 & {\scriptsize 0.41} & 39.4 & {\scriptsize 2.00} & 13.3$^{\dagger}$ & {\scriptsize 0.67} & 18.7 & {\scriptsize 0.39} & 64.1 & {\scriptsize 1.25} & 28.9$^{\dagger}$ & {\scriptsize 0.53} \\ 
& 1 & 9 & 09.1 & {\scriptsize 0.44} & 35.3 & {\scriptsize 2.39} & \textbf{14.5}$^{*}$ & {\scriptsize 0.76} & 18.9 & {\scriptsize 0.47} & 64.8 & {\scriptsize 1.12} & 29.3$^{\dagger}$ & {\scriptsize 0.63} \\

~~~~~~~~~~~~~~~+ hybrid
& 1 & 4 & 07.7 & {\scriptsize 0.17} & 40.0 & {\scriptsize 1.18} & 12.9$^{\dagger}$ & {\scriptsize 0.29} & 20.7 & {\scriptsize 0.59} & 61.1 & {\scriptsize 1.25} & \textbf{30.9}$^{\dagger}$ & {\scriptsize 0.71} \\
& 1 & 9 & 08.5 & {\scriptsize 0.32} & 36.6 & {\scriptsize 2.09} & 13.8$^{\dagger}$ & {\scriptsize 0.55} & 19.4 & {\scriptsize 0.36} & 62.9 & {\scriptsize 1.20} & 29.7$^{\dagger}$ & {\scriptsize 0.44} \\

~~~~~~~~~~~~w/ closest cluster selection
& 1 & 4 & 08.3 & {\scriptsize 0.25} & 40.8 & {\scriptsize 1.00} & 13.8$^{\dagger}$ & {\scriptsize 0.39} & 19.3 & {\scriptsize 0.21} & 63.4 & {\scriptsize 1.08} & 29.5$^{\dagger}$ & {\scriptsize 0.22} \\
& 1 & 9 & 08.7 & {\scriptsize 0.27} & 36.7 & {\scriptsize 1.12} & 14.1$^{\dagger}$ & {\scriptsize 0.43} & 19.8 & {\scriptsize 0.41} & 63.9 & {\scriptsize 0.70} & 30.2$^{\dagger}$ & {\scriptsize 0.50} \\

~~~~~~~~~~~~~~~+ hybrid
& 1 & 4 & 07.8 & {\scriptsize 0.20} & 41.6 & {\scriptsize 1.48} & 13.1$^{\dagger}$ & {\scriptsize 0.34} & 20.4 & {\scriptsize 0.47} & 61.0 & {\scriptsize 1.82} & 30.6$^{\dagger}$ & {\scriptsize 0.65} \\
& 1 & 9 & 08.6 & {\scriptsize 0.35} & 37.8 & {\scriptsize 1.52} & 14.0$^{\dagger}$ & {\scriptsize 0.55} & 19.2 & {\scriptsize 0.33} & 62.5 & {\scriptsize 1.23} & 29.4$^{\dagger}$ & {\scriptsize 0.45} \\

~~~~~~~~~and $k$-means++, \emph{summarization} \\
~~~~~~~~~~~~w/ furthest cluster selection
& 1 & 4 & 08.1 & {\scriptsize 0.26} & 38.0 & {\scriptsize 0.83} & 13.4$^{\dagger}$ & {\scriptsize 0.40} & 19.1 & {\scriptsize 0.33} & 64.2 & {\scriptsize 1.06} & 29.4$^{\dagger}$ & {\scriptsize 0.36} \\ 
& 1 & 9 & 09.1 & {\scriptsize 0.44} & 35.2 & {\scriptsize 1.61} & \textbf{14.5}$^{*}$ & {\scriptsize 0.66} & 19.1 & {\scriptsize 0.26} & 64.9 & {\scriptsize 1.09} & 29.5$^{\dagger}$ & {\scriptsize 0.29} \\ 

~~~~~~~~~~~~~~~+ hybrid
& 1 & 4 & 07.7 & {\scriptsize 0.16} & 40.3 & {\scriptsize 0.78} & 12.9$^{\dagger}$ & {\scriptsize 0.23} & 21.0 & {\scriptsize 0.40} & 61.4 & {\scriptsize 1.06} & \textbf{31.3}$^{\dagger}$ & {\scriptsize 0.38} \\
& 1 & 9 & 08.4 & {\scriptsize 0.29} & 36.6 & {\scriptsize 1.97} & 13.7$^{\dagger}$ & {\scriptsize 0.50} & 19.7 & {\scriptsize 0.29} & 63.3 & {\scriptsize 1.42} & 30.0$^{\dagger}$ & {\scriptsize 0.38} \\

~~~~~~~~~~~~w/ closest cluster selection
& 1 & 4 & 08.1 & {\scriptsize 0.21} & 38.6 & {\scriptsize 1.36} & 13.4$^{\dagger}$ & {\scriptsize 0.36} & 19.4 & {\scriptsize 0.17} & 62.9 & {\scriptsize 1.50} & 29.6$^{\dagger}$ & {\scriptsize 0.22} \\
& 1 & 9 & 08.9 & {\scriptsize 0.38} & 36.0 & {\scriptsize 1.81} & 14.3$^{\dagger}$ & {\scriptsize 0.63} & 19.2 & {\scriptsize 0.35} & 63.0 & {\scriptsize 1.04} & 29.4$^{\dagger}$ & {\scriptsize 0.43} \\ 

~~~~~~~~~~~~~~~+ hybrid
& 1 & 4 & 07.8 & {\scriptsize 0.20} & 40.2 & {\scriptsize 1.64} & 13.0$^{\dagger}$ & {\scriptsize 0.34} & 20.5 & {\scriptsize 0.46} & 60.7 & {\scriptsize 1.43} & 30.6$^{\dagger}$ & {\scriptsize 0.59} \\
& 1 & 9 & 08.7 & {\scriptsize 0.19} & 38.4 & {\scriptsize 1.37} & 14.2$^{\dagger}$ & {\scriptsize 0.31} & 19.2 & {\scriptsize 0.46} & 62.1 & {\scriptsize 1.76} & 29.3$^{\dagger}$ & {\scriptsize 0.65} \\ \bottomrule
                                \end{tabular}
\caption{
Full results obtained with \tacred{} using \gemmasmall{} and \gemmalarge{}.
This table complements Table \ref{t:results_fs-tacred_main} and \ref{t:results_fs-tacred_full_qwen}
by providing results with two additional LLMs. 
Incorporating additional examples results in minor improvements with \gemmasmall{}
and inconsistent gains with \gemmalarge.
$^{*}$ and $^{\dagger}$ indicate the best method in the whole column is statistically significantly better at $p < 0.05$ and $p < 0.01$, respectively, 
  based on one-sided paired bootstrap test with replacement on F1 scores.
}
  \label{t:results_fs-tacred_full_gemma}
\end{table*}

\begin{table*}
  \centering
  \small
  \setlength{\tabcolsep}{.061in}
  \setlength{\tabcolsep}{0.055in}
\begin{tabular}{l r@{+}l
                r@{ \scriptsize$\pm$ }l r@{ \scriptsize$\pm$ }l l@{ \scriptsize$\pm$ }l
                r@{ \scriptsize$\pm$ }l r@{ \scriptsize$\pm$ }l l@{ \scriptsize$\pm$ }l}
\toprule
  &
  \multicolumn{2}{c}{\#ex.} &
  \multicolumn{6}{c}{\gemmasmall} &
  \multicolumn{6}{c}{\gemmalarge}
  \\ \cmidrule(lr){4-9} \cmidrule(lr){10-15} 
& G & A &
  \multicolumn{2}{c}{P} & \multicolumn{2}{c}{R} & \multicolumn{2}{c}{F1} &
  \multicolumn{2}{c}{P} & \multicolumn{2}{c}{R} & \multicolumn{2}{c}{F1} \\ \midrule

Few-shot baseline & 1 & 0 & 35.1 & {\scriptsize 0.93} & 22.8$^{\dagger}$ & {\scriptsize 0.98} & 27.6 & {\scriptsize 0.96} & {36.7} & {\scriptsize {0.82}} & 60.3 & {\scriptsize 1.58} & {45.6} & {\scriptsize {0.81}} \\ \midrule
Including \emph{A}dditional examples \\

~~~Paraphrased with LLM       
& 1 & 4 & 32.4 & {\scriptsize 1.65} & 21.7 & {\scriptsize 0.92} & 26.0$^{\dagger}$ & {\scriptsize 1.12} & 36.5 & {\scriptsize 0.67} & 56.8 & {\scriptsize 1.52} & 44.4$^{\dagger}$ & {\scriptsize 0.70} \\ 
& 1 & 9 & 30.1 & {\scriptsize 0.89} & 17.7 & {\scriptsize 0.60} & 22.3$^{\dagger}$ & {\scriptsize 0.64} & 36.2 & {\scriptsize 0.67} & 59.5 & {\scriptsize 1.58} & \textbf{45.0}$^{\dagger}$ & {\scriptsize 0.69} \\

~~~Generated with LLM     
& 1 & 4 & 32.0 & {\scriptsize 0.84} & 43.9 & {\scriptsize 1.06} & \textbf{37.0} & {\scriptsize 0.60} & 36.5 & {\scriptsize 0.67} & 56.8 & {\scriptsize 1.52} & 44.4$^{\dagger}$ & {\scriptsize 0.70} \\ 
& 1 & 9 & 35.6 & {\scriptsize 1.21} & 37.6 & {\scriptsize 1.70} & 36.5$^{*}$ & {\scriptsize 1.20} & 36.2 & {\scriptsize 0.74} & 59.5 & {\scriptsize 1.48} & \textbf{45.0}$^{\dagger}$ & {\scriptsize 0.69} \\ \midrule

~~~Retrieved (closest) using \\

~~~~~~\sbert{} representations
& 1 & 4 & 35.6 & {\scriptsize 1.76} & 35.3 & {\scriptsize 1.85} & 35.4$^{\dagger}$ & {\scriptsize 1.69} & 34.6 & {\scriptsize 0.85} & 57.7 & {\scriptsize 1.39} & 43.2$^{\dagger}$ & {\scriptsize 0.87} \\  
& 1 & 9 & 37.3 & {\scriptsize 1.63} & 30.1 & {\scriptsize 1.54} & 33.3$^{\dagger}$ & {\scriptsize 1.35} & 36.0 & {\scriptsize 0.89} & 53.5 & {\scriptsize 1.58} & 43.0$^{\dagger}$ & {\scriptsize 0.97} \\

~~~~~~~~~+ hybrid
& 1 & 4 & 33.2 & {\scriptsize 0.70} & 42.7 & {\scriptsize 1.34} & \textbf{37.4} & {\scriptsize {0.65}} & 34.6 & {\scriptsize 1.04} & 58.2 & {\scriptsize 1.61} & \textbf{43.4}$^{\dagger}$ & {\scriptsize 1.03} \\
& 1 & 9 & {37.8} & {\scriptsize {1.13}} & 36.7 & {\scriptsize 1.30} & 37.2 & {\scriptsize 0.84} & 36.0 & {\scriptsize 0.78} & 54.6 & {\scriptsize 1.06} & 43.3$^{\dagger}$ & {\scriptsize 0.64} \\ \addlinespace

~~~~~~Semantic rule representations
& 1 & 4 & 27.8 & {\scriptsize 0.86} & 45.7 & {\scriptsize 1.83} & 34.6$^{\dagger}$ & {\scriptsize 0.95} & 31.5 & {\scriptsize 0.86} & 60.3 & {\scriptsize 1.61} & 41.4$^{\dagger}$ & {\scriptsize 0.95} \\ 
 & 1 & 9 & 28.4 & {\scriptsize 0.55} & 40.3 & {\scriptsize 1.57} & 33.3$^{\dagger}$ & {\scriptsize 0.80} & 31.4 & {\scriptsize 1.04} & 56.8 & {\scriptsize 1.41} & 40.4$^{\dagger}$ & {\scriptsize 1.05} \\

~~~~~~~~~+ hybrid 
 & 1 & 4 & 31.3 & {\scriptsize 0.78} & 45.3 & {\scriptsize 1.99} & \textbf{37.0} & {\scriptsize 1.07} & 31.7 & {\scriptsize 0.96} & 60.3 & {\scriptsize 1.70} & \textbf{41.5}$^{\dagger}$ & {\scriptsize 1.05} \\
 & 1 & 9 & 33.2 & {\scriptsize 0.70} & 39.5 & {\scriptsize 1.80} & 36.1$^{\dagger}$ & {\scriptsize 0.83} & 31.3 & {\scriptsize 1.15} & 56.8 & {\scriptsize 1.80} & 40.3$^{\dagger}$ & {\scriptsize 1.28} \\

~~~~~~~~~and $k$-means clustering \\
~~~~~~~~~~~~w/ random cluster selection
 & 1 & 4 & 27.6 & {\scriptsize 1.06} & 44.0 & {\scriptsize 2.07} & 33.9$^{\dagger}$ & {\scriptsize 1.27} & 31.3 & {\scriptsize 0.83} & 61.3 & {\scriptsize 1.75} & 41.4$^{\dagger}$ & {\scriptsize 0.97} \\ 
 & 1 & 9 & 26.6 & {\scriptsize 0.52} & 37.1 & {\scriptsize 1.12} & 31.0$^{\dagger}$ & {\scriptsize 0.66} & 30.4 & {\scriptsize 0.84} & 57.4 & {\scriptsize 1.68} & 39.7$^{\dagger}$ & {\scriptsize 0.87} \\

~~~~~~~~~~~~~~~+ hybrid 
 & 1 & 4 & 31.1 & {\scriptsize 0.57} & 45.0 & {\scriptsize 1.55} & \textbf{36.8}$^{*}$ & {\scriptsize 0.68} & 31.5 & {\scriptsize 0.84} & 61.3 & {\scriptsize 1.62} & \textbf{41.6}$^{\dagger}$ & {\scriptsize 0.92} \\
 & 1 & 9 & 33.6 & {\scriptsize 0.69} & 39.0 & {\scriptsize 1.78} & 36.1$^{\dagger}$ & {\scriptsize 1.00} & 30.5 & {\scriptsize 0.86} & 58.6 & {\scriptsize 1.46} & 40.1$^{\dagger}$ & {\scriptsize 0.89} \\

~~~~~~~~~~~~w/ closest cluster selection
& 1 & 4 & 27.3 & {\scriptsize 0.94} & 43.6 & {\scriptsize 1.43} & 33.6$^{\dagger}$ & {\scriptsize 0.98} & 31.0 & {\scriptsize 0.94} & 61.4 & {\scriptsize 2.28} & 41.2$^{\dagger}$ & {\scriptsize 1.20} \\
 & 1 & 9 & 27.5 & {\scriptsize 0.86} & 39.6 & {\scriptsize 0.96} & 32.5$^{\dagger}$ & {\scriptsize 0.84} & 30.1 & {\scriptsize 0.93} & 57.7 & {\scriptsize 1.87} & 39.6$^{\dagger}$ & {\scriptsize 1.12} \\

~~~~~~~~~~~~~~~+ hybrid
 & 1 & 4 & 30.8 & {\scriptsize 0.27} & 45.0 & {\scriptsize 1.28} & 36.6$^{*}$ & {\scriptsize 0.42} & 31.2 & {\scriptsize 0.73} & 61.9 & {\scriptsize 2.13} & 41.5$^{\dagger}$ & {\scriptsize 0.93} \\
 & 1 & 9 & 33.0 & {\scriptsize 0.67} & 38.6 & {\scriptsize 1.78} & 35.5$^{\dagger}$ & {\scriptsize 1.03} & 30.6 & {\scriptsize 1.10} & 59.0 & {\scriptsize 1.90} & 40.3$^{\dagger}$ & {\scriptsize 1.25} \\

~~~~~~~~~and $k$-means++ clustering \\
~~~~~~~~~~~~w/ furthest cluster selection
 & 1 & 4 & 28.1 & {\scriptsize 0.55} & 45.5 & {\scriptsize 1.49} & 34.8$^{\dagger}$ & {\scriptsize 0.65} & 31.2 & {\scriptsize 0.55} & 60.5 & {\scriptsize 1.68} & 41.1$^{\dagger}$ & {\scriptsize 0.69} \\ 
 & 1 & 9 & 25.3 & {\scriptsize 0.89} & 33.4 & {\scriptsize 1.02} & 28.8$^{\dagger}$ & {\scriptsize 0.76} & 30.4 & {\scriptsize 0.81} & 57.6 & {\scriptsize 1.60} & 39.7$^{\dagger}$ & {\scriptsize 0.88} \\ 

~~~~~~~~~~~~~~~+ hybrid 
 & 1 & 4 & 31.1 & {\scriptsize 0.48} & 44.8 & {\scriptsize 1.74} & 36.7$^{*}$ & {\scriptsize 0.58} & 31.7 & {\scriptsize 0.60} & 60.9 & {\scriptsize 1.58} & \textbf{41.7}$^{\dagger}$ & {\scriptsize 0.71} \\
 & 1 & 9 & 33.5 & {\scriptsize 0.72} & 38.2 & {\scriptsize 1.75} & 35.6$^{\dagger}$ & {\scriptsize 1.10} & 30.4 & {\scriptsize 0.76} & 58.6 & {\scriptsize 1.79} & 40.1$^{\dagger}$ & {\scriptsize 0.85} \\

~~~~~~~~~~~~w/ closest cluster selection
& 1 & 4 & 28.6 & {\scriptsize 0.80} & 47.1 & {\scriptsize 1.78} & 35.6$^{\dagger}$ & {\scriptsize 0.94} & 31.3 & {\scriptsize 0.89} & 61.3 & {\scriptsize 2.06} & 41.4$^{\dagger}$ & {\scriptsize 1.00} \\
 & 1 & 9 & 27.1 & {\scriptsize 0.66} & 38.6 & {\scriptsize 1.68} & 31.8$^{\dagger}$ & {\scriptsize 0.89} & 30.3 & {\scriptsize 1.05} & 57.9 & {\scriptsize 1.73} & 39.8$^{\dagger}$ & {\scriptsize 1.17} \\ 

~~~~~~~~~~~~~~~+ hybrid 
 & 1 & 4 & 31.1 & {\scriptsize 1.00} & 45.4 & {\scriptsize 1.36} & \textbf{36.9} & {\scriptsize 0.94} & 31.4 & {\scriptsize 0.79} & 61.5 & {\scriptsize 1.82} & 41.6$^{\dagger}$ & {\scriptsize 0.87} \\
 & 1 & 9 & 33.8 & {\scriptsize 0.71} & 39.4 & {\scriptsize 1.52} & 36.4$^{\dagger}$ & {\scriptsize 0.82} & 30.2 & {\scriptsize 1.12} & 58.1 & {\scriptsize 1.80} & 39.7$^{\dagger}$ & {\scriptsize 1.23} \\ \midrule

~~~Retrieved (closest) using \\
~~~~~~Semantic rule representations \\
~~~~~~~~~and $k$-means, \emph{summarization} \\
~~~~~~~~~~~~w/ random cluster selection
 & 1 & 4 & 27.6 & {\scriptsize 1.18} & 47.9 & {\scriptsize 1.79} & 35.0$^{\dagger}$ & {\scriptsize 1.18} & 31.6 & {\scriptsize 1.04} & 62.1 & {\scriptsize 1.30} & 41.9$^{\dagger}$ & {\scriptsize 1.08} \\ 
 & 1 & 9 & 28.0 & {\scriptsize 0.65} & 41.2 & {\scriptsize 0.95} & 33.3$^{\dagger}$ & {\scriptsize 0.49} & 31.3 & {\scriptsize 0.95} & 60.2 & {\scriptsize 1.63} & 41.2$^{\dagger}$ & {\scriptsize 1.03} \\

~~~~~~~~~~~~~~~+ hybrid 
 & 1 & 4 & 28.7 & {\scriptsize 0.30} & 48.2 & {\scriptsize 1.64} & \textbf{36.0}$^{\dagger}$ & {\scriptsize 0.42} & 31.7 & {\scriptsize 0.82} & 61.9 & {\scriptsize 1.04} & 41.9$^{\dagger}$ & {\scriptsize 0.77} \\
 & 1 & 9 & 30.0 & {\scriptsize 0.47} & 44.0 & {\scriptsize 1.75} & 35.6$^{\dagger}$ & {\scriptsize 0.65} & 31.0 & {\scriptsize 0.85} & 60.6 & {\scriptsize 1.56} & 41.0$^{\dagger}$ & {\scriptsize 0.90} \\

~~~~~~~~~~~~w/ closest cluster selection
 & 1 & 4 & 26.8 & {\scriptsize 0.74} & 47.5 & {\scriptsize 1.78} & 34.2$^{\dagger}$ & {\scriptsize 0.83} & 31.7 & {\scriptsize 0.74} & {62.2} & {\scriptsize {1.45}} & \textbf{42.0}$^{\dagger}$ & {\scriptsize 0.76} \\
 & 1 & 9 & 27.7 & {\scriptsize 0.49} & 42.7 & {\scriptsize 1.02} & 33.6$^{\dagger}$ & {\scriptsize 0.40} & 31.7 & {\scriptsize 0.82} & 60.7 & {\scriptsize 1.60} & 41.6$^{\dagger}$ & {\scriptsize 0.81} \\

~~~~~~~~~~~~~~~+ hybrid 
 & 1 & 4 & 28.5 & {\scriptsize 0.30} & 48.2 & {\scriptsize 1.62} & 35.8$^{\dagger}$ & {\scriptsize 0.51} & 31.5 & {\scriptsize 0.84} & 61.9 & {\scriptsize 1.61} & 41.8$^{\dagger}$ & {\scriptsize 0.89} \\
 & 1 & 9 & 29.7 & {\scriptsize 0.53} & 44.2 & {\scriptsize 1.35} & 35.5$^{\dagger}$ & {\scriptsize 0.59} & 31.1 & {\scriptsize 0.90} & 60.5 & {\scriptsize 1.43} & 41.0$^{\dagger}$ & {\scriptsize 0.90} \\

~~~~~~~~~and $k$-means++, \emph{summarization} \\
~~~~~~~~~~~~w/ furthest cluster selection
 & 1 & 4 & 28.0 & {\scriptsize 0.71} & 45.7 & {\scriptsize 1.47} & 34.7$^{\dagger}$ & {\scriptsize 0.71} & 31.4 & {\scriptsize 0.71} & 62.0 & {\scriptsize 1.58} & 41.7$^{\dagger}$ & {\scriptsize 0.79} \\ 
 & 1 & 9 & 28.0 & {\scriptsize 0.84} & 40.1 & {\scriptsize 1.52} & 33.0$^{\dagger}$ & {\scriptsize 0.91} & 31.3 & {\scriptsize 0.86} & 59.7 & {\scriptsize 1.36} & 41.1$^{\dagger}$ & {\scriptsize 0.85} \\ 

~~~~~~~~~~~~~~~+ hybrid 
 & 1 & 4 & 28.9 & {\scriptsize 0.37} & 47.4 & {\scriptsize 1.77} & 35.9$^{\dagger}$ & {\scriptsize 0.70} & 31.5 & {\scriptsize 0.86} & 61.9 & {\scriptsize 1.70} & 41.8$^{\dagger}$ & {\scriptsize 0.90} \\
 & 1 & 9 & 30.5 & {\scriptsize 0.60} & 43.7 & {\scriptsize 1.40} & 35.9$^{\dagger}$ & {\scriptsize 0.66} & 30.9 & {\scriptsize 0.87} & 60.0 & {\scriptsize 1.49} & 40.7$^{\dagger}$ & {\scriptsize 0.81} \\

~~~~~~~~~~~~w/ closest cluster selection
 & 1 & 4 & 27.8 & {\scriptsize 0.67} & {48.5} & {\scriptsize {1.39}} & 35.3$^{\dagger}$ & {\scriptsize 0.65} & 31.7 & {\scriptsize 1.02} & 62.1 & {\scriptsize 1.92} & 42.0$^{\dagger}$ & {\scriptsize 1.17} \\
 & 1 & 9 & 28.7 & {\scriptsize 0.63} & 44.5 & {\scriptsize 1.38} & 34.9$^{\dagger}$ & {\scriptsize 0.74} & 31.9 & {\scriptsize 0.97} & 60.2 & {\scriptsize 1.51} & 41.7$^{\dagger}$ & {\scriptsize 0.99} \\ 

~~~~~~~~~~~~~~~+ hybrid 
 & 1 & 4 & 28.9 & {\scriptsize 0.62} & 48.4 & {\scriptsize 1.92} & \textbf{36.2}$^{\dagger}$ & {\scriptsize 0.91} & 31.9 & {\scriptsize 0.84} & 61.9 & {\scriptsize 1.55} & \textbf{42.1}$^{\dagger}$ & {\scriptsize 0.87} \\
 & 1 & 9 & 30.3 & {\scriptsize 0.61} & 43.9 & {\scriptsize 1.40} & 35.9$^{\dagger}$ & {\scriptsize 0.71} & 31.2 & {\scriptsize 0.98} & 60.3 & {\scriptsize 1.54} & 41.1$^{\dagger}$ & {\scriptsize 1.03} \\ \bottomrule

\end{tabular}
  \caption{
  Full results obtained with \fewrel{} using \gemmasmall{} and \gemmalarge{}.
This table complements Table \ref{t:results_fs-fewrel_main} and \ref{t:results_fs-fewrel_full_qwen}
by providing results with two additional LLMs. 
Incorporating additional examples results in minor improvements with \gemmasmall{},
and inconsistent gains with \gemmalarge{}---the $1$-shot baseline is the best-performing system.
$^{*}$ and $^{\dagger}$ indicate the best method in the whole column is statistically significantly better at $p < 0.05$ and $p < 0.01$, respectively, 
  based on one-sided paired bootstrap test with replacement on F1 scores.
}
  \label{t:results_fs-fewrel_full_gemma}
\end{table*}

\begin{table*}
\centering
\small
\begin{tabular}{l c rr rr}
\toprule
& \multirow{2}{*}{\# Addtl. examples}
& \multicolumn{2}{c}{\tacred}
& \multicolumn{2}{c}{\fewrel} \\ \cmidrule(lr){3-4} \cmidrule(lr){5-6}

&& \# Generated & \# Retrieved & \# Generated & \# Retrieved \\ \midrule

\sbert{} representations
& 4 & 2.94 & 1.06 & 3.03 & 0.97 \\
& 9 & 7.10 & 2.35 & 7.16 & 1.84 \\ \addlinespace

Semantic rule representations
& 4 & 2.77 & 1.23 & 2.88 & 1.12 \\
& 9 & 6.65 & 2.35 & 6.66 & 2.34 \\

~~~and $k$-means clustering 
& 4 & 3.05 & 0.95 & 3.08 & 0.92 \\
& 9 & 7.18 & 1.82 & 7.11 & 1.89 \\

\bottomrule
\end{tabular}

\caption{
Analysis of our hybrid approach using Qwen3-4B on \tacred{} and \fewrel{} (best performing methods in Tables \ref{t:results_fs-tacred_main} and \ref{t:results_fs-fewrel_main}).
Regardless of whether we aim for 4 or 9 additional examples and the dataset,
the majority (roughly 2/3) of additional examples are selected from the ones generated with an LLM.
The remaining one third, however, are selected from our retrieval approach
grounded using semantic rule representations and clustering.
Recall that the hybrid approach (as opposed to using only examples generated with an LLM) yields better results (Tables \ref{t:results_fs-tacred_main} and \ref{t:results_fs-fewrel_main}).
We hypothesize that this is due to the higher diversity in the retrieved examples (Section \ref{sec:qualitative_analysis}).
}
\label{t:results_fs-tacred_example_picks}
\end{table*}

\begin{table*}
  \centering
  \small
  \begin{tabular}{l r@{+}l rr rr}
\toprule
\multirow{2}{*}{Method}
  & \multicolumn{2}{c}{\#ex}
  & \multicolumn{2}{c}{Gold. vs. Addtl.}
  & \multicolumn{2}{c}{Among Addtl.} \\ \cmidrule(lr){4-5} \cmidrule(lr){6-7}

  & G &  A& \% overlap & cosine & \% overlap & cosine \\ \midrule

Paraphrased with LLM & 1 & 4 &  $41.0$ & $0.827$ & $54.1$ & $0.867$ \\ 
                     & 1 & 9 &  $41.3$ & $0.827$ & $59.1$ & $0.880$ \\   
Generated with LLM   & 1 & 4 &  $ 7.6$ & $0.275$ & $12.9$ & $0.379$ \\ 
                     & 1 & 9 &  $ 8.0$ & $0.272$ & $17.0$ & $0.398$ \\
\midrule
Retrieved (closest) using \\
~~~\sbert{} representations        & 1 & 4 &  $ 8.3$ & $0.361$ &  $08.9$ & $0.339$\\ 
                              & 1 & 9 &  $ 8.6$ & $0.345$ &  $08.2$ & $0.311$\\ \addlinespace
                              
~~~Semantic rule representations & 1 & 4 &  $12.8$ & $0.257$ &  $12.3$ & $0.251$\\ 
                                 & 1 & 9 &  $11.8$ & $0.245$ &  $11.0$ & $0.239$\\
~~~~~~+ hybrid                   & 1 & 4 &  $ 9.4$ & $0.267$ &  $09.9$ & $0.303$\\ 
                                 & 1 & 9 &  $ 9.2$ & $0.269$ &  $12.6$ & $0.326$\\
~~~and $k$-means clustering        & 1 & 4 &  $ 7.3$ & $0.171$ &  $ 6.8$ & $0.158$\\ 
                                 & 1 & 9 &  $ 7.4$ & $0.170$ &  $ 6.9$ & $0.160$\\
~~~~~~~+ hybrid                  & 1 & 4 &  $ 6.8$ & $0.278$ &  $ 9.8$ & $0.308$\\ 
                                 & 1 & 9 &  $ 7.2$ & $0.247$ &  $12.8$ & $0.367$\\\bottomrule

\end{tabular}
\caption{
Diversity analysis of all the additional examples obtained with several methods on \tacred.
We report the average diversity metrics between
(a) the gold support sentence (i.e., original example) and the additional examples (columns 3--4)
and
(b) among the additional examples (columns 5--6).
The diversity metrics are the percentage of tokens overlap
and the cosine similarity of the SBERT representations.
Lower values indicate higher diversity with both metrics.
Paraphrasing with an LLM results in additional examples with high token overlap and cosine similarities, indicating low diversity.
Generating with an LLM alleviates the issue, but retrieving with semantic rule representations results in additional examples with the lowest token overlaps and cosine similarity.
Note that combining both of these examples (hybrid) is more beneficial for in-context learning (Table \ref{t:results_fs-tacred_full_qwen}),
thus the relation at hand does hold despite the higher diversity.
}
  \label{t:diversity_comparison}
\end{table*}

\begin{table*}
  \centering
  \small
  \setlength{\tabcolsep}{.061in}
\begin{tabular}{l r@{+}l
                r@{ \scriptsize $\pm$ }l r@{ \scriptsize $\pm$ }l r@{ \scriptsize $\pm$ }l
                r@{ \scriptsize $\pm$ }l r@{ \scriptsize $\pm$ }l r@{ \scriptsize $\pm$ }l}
\toprule 
  Dataset &
  \multicolumn{2}{c}{\#ex.} &
  \multicolumn{6}{c}{\qwensmall} &
  \multicolumn{6}{c}{\gemmasmall} \\ \cmidrule(lr){4-9} \cmidrule(lr){10-15}
& G & A &
  \multicolumn{2}{c}{P} & \multicolumn{2}{c}{R} & \multicolumn{2}{c}{F1} &
  \multicolumn{2}{c}{P} & \multicolumn{2}{c}{R} & \multicolumn{2}{c}{F1}\\ \midrule

\tacred{} \\
~~~Binary    & 1 & 0 & 28.7 & {\scriptsize 0.99} & 17.8 & {\scriptsize 1.21} & \textbf{22.0} & {\scriptsize {1.19}} & {11.9} & {\scriptsize {0.60}} & 18.5 & {\scriptsize 0.54} & \textbf{14.5} & {\scriptsize {0.58}}  \\

~~~Multi-class    & 1 & 0 & 05.0 & {\scriptsize 0.15} & 64.4 & {\scriptsize 1.11} & 09.3 & {\scriptsize 0.26} & 05.5 & {\scriptsize 0.18} & {60.5} & {\scriptsize {1.38}} & 10.2 & {\scriptsize 0.31}  \\ \midrule

\fewrel{} \\
~~~Binary   & 1 & 0 & 41.7 & {\scriptsize 1.37} & 26.4 & {\scriptsize 1.17} & \textbf{32.4} & {\scriptsize {1.14}} & {35.1} & {\scriptsize {0.93}} & 22.8 & {\scriptsize 0.98} & 27.6 & {\scriptsize 0.96}  \\

~~~Multi-class    & 1 & 0 & 14.5 & {\scriptsize 0.59} & 53.8 & {\scriptsize 1.81} & 22.9 & {\scriptsize 0.86} & 23.9 & {\scriptsize 0.47} & {59.3} & {\scriptsize {1.87}} & \textbf{34.0} & {\scriptsize {0.62}}  \\

\bottomrule
\end{tabular}
\caption{
Results with the $1$-shot setting using the
binary (for each of the five relations in an episode, does it hold between the entities at hand?)
and 
multi-class approaches (which of these five relations holds between the entities at hand, if any?).
See descriptions of both approaches in Section \ref{sec:few-shot-relation-extraction-task}.
The binary approach yields stronger results for all corpora and LLMs except \gemmasmall{} with \fewrel{}, so we report all our main results with the binary approach.
}

  \label{t:results_all_relation_single_prompt}
\end{table*}

\begin{table*}
  \centering
  \small
  \setlength{\tabcolsep}{.061in}
\begin{tabular}{l r@{+}l
                r@{ \scriptsize $\pm$ }l r@{ \scriptsize $\pm$ }l r@{ \scriptsize $\pm$ }l
                r@{ \scriptsize $\pm$ }l r@{ \scriptsize $\pm$ }l r@{ \scriptsize $\pm$ }l}
\toprule 
  &
  \multicolumn{2}{c}{\#ex.} &
  \multicolumn{6}{c}{\qwensmall} &
  \multicolumn{6}{c}{\gemmasmall} \\ \cmidrule(lr){4-9} \cmidrule(lr){10-15}
& G & A &
  \multicolumn{2}{c}{P} & \multicolumn{2}{c}{R} & \multicolumn{2}{c}{F1} &
  \multicolumn{2}{c}{P} & \multicolumn{2}{c}{R} & \multicolumn{2}{c}{F1}\\ \midrule

\tacred                             & 1 & 0 & 28.7 & {\scriptsize 0.99} & 17.8 & {\scriptsize 1.21} & \textbf{22.0} & {\scriptsize {1.19}} & 11.9 & {\scriptsize 0.60} & 18.5 & {\scriptsize 0.54} & \textbf{14.5} & {\scriptsize {0.58}}  \\
~~~without NER filter              & 1 & 0 & 21.7 & {\scriptsize 0.65} & 18.9 & {\scriptsize 1.15} & 20.2 & {\scriptsize 0.93} & 07.6 & {\scriptsize 0.39} & 18.4 & {\scriptsize 0.56} & 10.8 & {\scriptsize 0.47} \\ \midrule
\fewrel                  & 1 & 0 & 41.7 & {\scriptsize 1.37} & 26.4 & {\scriptsize 1.17} & \textbf{32.4} & {\scriptsize {1.14}} & 35.1 & {\scriptsize 0.93} & 22.8 & {\scriptsize 0.98} & \textbf{27.6} & {\scriptsize {0.96}}  \\
~~~without NER filter   & 1 & 0 & 29.4 & {\scriptsize 1.14} & 28.1 & {\scriptsize 1.14} & 28.7 & {\scriptsize 0.98} & 20.9 & {\scriptsize 0.31} & 23.5 & {\scriptsize 0.92} & 22.1 & {\scriptsize 0.56} \\ 

\bottomrule
\end{tabular}
  \caption{
Ablation study on the effect of NER-based filtering in the baseline $1$-shot setting on \tacred{} and \fewrel{} using \qwensmall{} and \gemmasmall{} (first rows in Tables \ref{t:results_fs-tacred_main} and \ref{t:results_fs-fewrel_main}).
Recall that the deterministic filtering simply discards a relation at inference time unless the named entities of the subject and object named entities have the expected types.
Removing the filter substantially decreases precision while minimally decreasing recall (resulting in lower F1 across corpora and LLMs),
so we keep the filter in our main experiments.
}
  \label{t:results_without_ner_filter}
\end{table*}

\end{document}